\documentclass{article}

\PassOptionsToPackage{square,sort,comma,numbers}{natbib}

\usepackage[final]{neurips_2022}

\usepackage[utf8]{inputenc} % allow utf-8 input
\usepackage[T1]{fontenc}    % use 8-bit T1 fonts
\usepackage{hyperref}       % hyperlinks
\usepackage{url}            % simple URL typesetting
\usepackage{booktabs}       % professional-quality tables
\usepackage{amsfonts}       % blackboard math symbols
\usepackage{nicefrac}       % compact symbols for 1/2, etc.
\usepackage{microtype}      % microtypography
\usepackage{xcolor}         % colors

\usepackage{float}
\usepackage{wrapfig}
\newcommand{\mathbbm}[1]{\text{\usefont{U}{bbm}{m}{n}#1}}
\usepackage{amsmath}
\usepackage{amssymb}
\usepackage{mathtools}
\usepackage{amsthm}
\usepackage{bm}
\usepackage{mathtools}
\usepackage{stmaryrd}
\usepackage{dirtytalk}
\usepackage{makecell}
\usepackage{multirow}
\usepackage{csvsimple}
\usepackage{siunitx}
\usepackage{subcaption}

\usepackage{cleveref}
\Crefformat{figure}{#2Fig.~#1#3}
\Crefformat{theorem}{#2Thm.~#1#3}
\Crefformat{definition}{#2Def.~#1#3}
\Crefformat{algorithm}{#2Alg.~#1#3}
\Crefformat{section}{#2Sec.~#1#3}
\Crefmultiformat{section}{Secs.~#2#1#3}{ and~#2#1#3}{, #2#1#3}{ and~#2#1#3}

\theoremstyle{plain}
\newtheorem{theorem}{Theorem}[section]

\theoremstyle{definition}
\newtheorem{definition}[theorem]{Definition}

\theoremstyle{remark}

\title{Neural Stochastic PDEs: Resolution-Invariant Learning of Continuous Spatiotemporal Dynamics}

\author{
  Cristopher Salvi \\
  Imperial College London \& \\
  The Alan Turing Institute \\
  \texttt{c.salvi@imperial.ac.uk}

  \And
  Maud Lemercier \\
  University of Warwick \\
  \texttt{maud.lemercier@warwick.ac.uk}
  
  \And
  Andris Gerasimovi\v{c}s \\
  University of Bath \\
  \texttt{ag2616@bath.ac.uk} \\
 
}

\begin{document}

\maketitle

\begin{abstract}
   \emph{Stochastic partial differential equations} (SPDEs) are the mathematical tool of choice for modelling spatiotemporal PDE-dynamics under the influence of randomness. Based on the notion of mild solution of an SPDE, we introduce a novel neural architecture to learn solution operators of PDEs with (possibly stochastic) forcing from partially observed data. The proposed \emph{Neural SPDE} model provides an extension to two popular classes of physics-inspired architectures. On the one hand, it extends Neural CDEs and variants -- continuous-time analogues of RNNs -- in that it is capable of processing incoming sequential information arriving at arbitrary spatial resolutions. On the other hand, it extends Neural Operators -- generalizations of neural networks to model mappings between spaces of functions -- in that it can parameterize solution operators of SPDEs depending simultaneously on the initial condition and a realization of the driving noise. By performing operations in the spectral domain, we show how a Neural SPDE can be evaluated in two ways, either by calling an ODE solver (emulating a spectral Galerkin scheme), or by solving a fixed point problem. Experiments on various semilinear SPDEs, including the stochastic Navier-Stokes equations, demonstrate how the Neural SPDE model is capable of learning complex spatiotemporal dynamics in a resolution-invariant way, with better accuracy and lighter training data requirements compared to alternative models, and up to 3 orders of magnitude faster than traditional solvers.
\end{abstract}

\section{Introduction}\label{sec:intro}

\emph{Stochastic partial differential equations} (SPDEs) are the mathematical formalism used to model many physical, biological and economic systems subject to the influence of randomness, be it intrinsic (e.g. quantifying uncertainty) or extrinsic (e.g. modelling environmental random perturbations). Notable examples of SPDEs include the \emph{Kardar–Parisi–Zhang (KPZ) equation} for modelling random interface growth such as the propagation of a forest fire from a burnt region to an unburnt region \citep{hairer2013solving}, the \emph{Ginzburg-Landau equation} describing phase transitions of ferromagnets and superconductors near critical temperature \citep{temam2012infinite}, or the \emph{stochastic Navier-Stokes equations} modelling the dynamics of a turbulent fluid flow under the presence of local random fluctuations \citep{mikulevicius2004stochastic}. For an introduction to the theory of SPDEs see \citet{hairer2009introduction}; a comprehensive textbook is \citet{ holden1996stochastic}. 

Classical numerical approaches for solving SPDEs include finite difference methods and spectral Galerkin methods \citep{lord2014introduction} among others. To ensure accuracy and stability of numerical solutions of complex SPDEs, computations must be carried out at high resolution using fine discretization grids, rendering the resulting schemes computationally intractable. This limitation motivates the study of data-driven methods that can learn solutions to differential equations from partially observed data. 

\paragraph{Related work} There has been an increased interest in recent years to combine neural networks and differential equations into a hybrid approach \cite{weinan2017proposal, chen2018neural, kidger2022neural}. 

\emph{Neural controlled differential equations} (Neural CDEs), as popularised by  \cite{kidger2020neural,morrill2021neural,bellot2021policy}, are continuous-time analogues to recurrent neural networks (RNN, GRU, LSTM etc.). The input to a Neural CDE model is a multivariate time series interpolated into a continuous path $X: [0,T] \to \mathbb{R}^{d_\xi}$; the model consists of a matrix-valued feedforward neural network $f_\theta : \mathbb{R}^{d_h} \to \mathbb{R}^{d_h \times d_\xi}$ parameterizing the vector field of the following dynamical system (and satisfying some minimal Lipschitz regularity to ensure existence and uniqueness of solutions)
\begin{equation}\label{eqn:ncde}
    z_0 = \ell_\theta(u_0), \quad
    z_t = z_0 + \int_0^tf_\theta(z_s)dX_s, \quad
    u_t = \pi_\theta(z_t),
\end{equation}
where $\ell_\theta:\mathbb{R}^{d_u} \to \mathbb{R}^{d_h}$ and $\pi_\theta:\mathbb{R}^{d_h} \to \mathbb{R}^{d_u}$ are feedforward neural networks. The output response $u : [0,T] \to \mathbb{R}^{d_u}$ is then fed to a (possibly pathwise) loss function (mean squared, cross entropy etc.) and trained via stochastic gradient descent in the usual way. In practice, the term \say{$dX_t$} means that the solution $z_t$ of the equation (e.g. attention required from a doctor) can change in response to a change of an external stream of information $X_t$ (e.g. heart rate of a patient). 

Depending on the level of roughness of the control path $X$, the integral in \cref{eqn:ncde} can be interpreted in different ways. In \citep{kidger2020neural}, $X$ is assumed differentiable and is obtained in practice via cubic splines interpolation of the original time series. In this way, the term \say{$dX_s$} can be interpreted as \say{$\dot X_s ds$} so that \cref{eqn:ncde} becomes an ODE of the form $\dot z_t = f_\theta(z_t)\dot X_t$ that can be evaluated numerically via a call to an ODE solver of choice (Euler, Runge-Kutta, implicit, adaptive stepsize schemes etc.). More generally, if $X$ is of bounded variation then the integral above can be seen as a classical Riemann–Stieltjes or Young integral \cite{young1905vi}. Neural SDEs \citep{ liu2019neural, li2020scalable, kidger2021neural, kidger2021efficient} are a special subclass of Neural CDEs where the control is a  sample path from a $d_\xi$-dimensional Brownian motion (which is not of bounded variation), and \cref{eqn:ncde} is understood via stochastic integration (It\^o, Stratonovich etc.). Neural RDEs \citep{morrill2021neural} allow to relax even further the regularity assumptions on $X$ by treating the integral using rough integration \cite{lyons1998differential, gubinelli2004controlling}. In practice, Neural RDEs are particularly well suited for long time series. This is due to the fact that the model can be evaluated via a numerical scheme from stochastic analysis (called the \emph{log-ODE method} \cite{morrill2021neural}) over intervals much larger than what would be expected given the sampling rate of the time series. However, the space complexity of the numerical solver increases exponentially in the number of channels $d_\xi$, thus model complexity becomes intractable for high dimensional time series. Despite offering many advantages for modelling temporal dynamics, these models are not designed to process signals varying both in space and in time such as physical fields described by SPDEs. In particular, although these models are time-resolution invariant, they are not space-resolution invariant, and are not well suited to capture nonlinear interactions between the various space-time points typically observed in SPDE-dynamics. Similar to CDEs, the solution $u$ to an SPDE is characterized by an initial condition $u_0$ and a driving noise $X$. However, in the case of CDEs, $(u_0,X_t,u_t)$ are vectors, while in the case of SPDEs they are functions. 

\emph{Neural Operators} \citep{kovachki2021neural, li2020multipole, li2020neural,lu2021learning} are generalizations of neural networks capable of modelling mappings between spaces of functions and offer an attractive option for learning with spatiotemporal data \citep{kovachki2021neural}. Among all kinds of Neural Operators, \emph{Fourier Neural Operators} (FNOs) \cite{li2020fourier} stand out because of their easier parametrization while demonstrating similar learning performance compared to other Neural Operator models. However, Neural Operators generally fail to incorporate the effect that an external (possibly random) spatiotemporal signal might have on the system they describe. In the case of SPDEs, the external signal is indeed random (e.g. sample from a Wiener process) and its presence leads to new phenomena, both at the mathematical and the physical level, often describing more complex and realistic dynamics than the ones arising from deterministic PDEs.

\paragraph{Contributions} To overcome the above limitations faced by Neural CDEs and Neural Operators, we introduce the \emph{neural stochastic partial differential equation} (Neural SPDE) model, capable of learning solution operators of SPDEs from partially observed data by processing, continuously in time and space, incoming sequential information arriving at an arbitrary resolution. We propose two separate algorithms to evaluate our model: the first reduces the Neural SPDE to a system of ODEs in Fourier space, which can then be solved numerically by means of any ODE solver of choice (emulating a spectral Galerkin scheme); the second rewrites the Neural SPDE as a fixed point problem, which is solved via classical root-finding schemes. For both choices of evaluation, the Neural SPDE model inherits memory-efficient backpropagation capabilities provided by existing adjoint-based and implicit-differentiation-based methods respectively. Finally, we perform extensive experiments on various semilinear SPDEs, including the stochastic Ginzburg-Landau, Korteweg-De Vries, Navier-Stokes equations. The empirical results illustrate several useful aspects of our model: 1) it is space and time resolution-invariant, meaning that even if trained on a lower resolution it can be directly evaluated on a higher resolution; 2) it requires a lower amount of training data to achieve similar or better performance compared to alternative models; 3) its evaluation is up to 3 orders of magnitude faster than traditional numerical solvers.

% Our code is available at {\small \url{ https://github.com/crispitagorico/Neural-SPDEs}}.

The outline of the paper is as follows: in \Cref{sec:background} we provide a brief introduction to SPDEs which will help us to define our Neural SPDE model in \Cref{sec:model}, followed by numerical experiments in \Cref{sec:experiments}. In \Cref{sec:comp_spdes}, we provide an overview of the computational aspects of SPDEs used to design the model and solve SPDEs numerically. Additional experiments can be found in \Cref{sec:experiments_appendix}.  

\section{Background on SPDEs}\label{sec:background}
% For simplicity, we will assume that all functions discussed in this section are complex valued. 
Let $T > 0$ and $d,d_u,d_\xi \in \mathbb{N}$. Let $\mathcal{D} \subset \mathbb{R}^d$ be a bounded domain. Let $\mathcal{H}_u=\{f:\mathcal{D}\to \mathbb{R}^{d_u}\}$ and $\mathcal{H}_\xi=\{f:\mathcal{D}\to \mathbb{R}^{d_\xi}\}$ be two Hilbert spaces of functions from $\mathcal{D}$ to $\mathbb{R}^{d_u}$ and $\mathbb{R}^{d_\xi}$ respectively. We consider a large class of SPDEs of the following type 
\begin{equation}\label{eqn:SPDE}
    du_t = \left(\mathcal{L}u_t + F(u_t)\right)dt + G(u_t)dW_t,
\end{equation}
where $W_t$ is either an infinite dimensional $Q$-Wiener process \citep[Def. 10.6]{lord2014introduction} or a cylindrical Wiener process \citep[Def. 3.54]{hairer2009introduction} with values in $\mathcal{H}_{\xi}$, $F : \mathcal{H}_u \to \mathcal{H}_u$ and $G : \mathcal{H}_u \to L(\mathcal{H}_\xi,\mathcal{H}_u)$ are two continuous operators, $L(\mathcal{H}_\xi,\mathcal{H}_u)$ is the space of bounded linear operators from $\mathcal{H}_\xi$ to $\mathcal{H}_u$, and $\mathcal{L}$ is a linear differential operator generating a \emph{semigroup}\footnote{A strongly continuous semigroup $S$ on $\mathcal{H}_u$ is a family of bounded linear operators $S = \left\{S_t : \mathcal{H}_u \to \mathcal{H}_u\right\}_{t\geq0}$ with the properties that: 1) $S_0=\text{Id}$, the identity operator on $\mathcal{H}_u$, 
2) $S_t \circ S_s = S_{t+s}$, for any $s,t \geq 0$, and 3) the function $t \mapsto S_tu$ is continuous from $[0,T]$ to $\mathcal{H}_u$, for any $u \in \mathcal{H}_u$.} $e^{t\mathcal{L}} : \mathcal{H}_u \to \mathcal{H}_u$. For further details on Wiener processes see \Cref{ssec:wiener}, and for a primer on semigroup theory see \citet[Section 4]{hairer2009introduction}. A function $u:[0,T]\to\mathcal{H}_u$ is said to be a \emph{mild solution} of the SPDE (\ref{eqn:SPDE}) if for any $t \in [0,T]$ it satisfies
\begin{align*}
    u_t &= e^{t\mathcal{L}}u_0 + \int_0^te^{(t-s)\mathcal{L}}F(u_s)ds + \int_0^te^{(t-s)\mathcal{L}}G(u_s)dW_s,
\end{align*}
where the second integral is a stochastic integral interpreted in the It\^o sense \citep[Def. 3.57]{hairer2009introduction}. Thus, an SPDE can be informally thought of as an SDE with values in the functional space $\mathcal{H}_u$ and driven by an infinite dimensional Brownian motion $W$. Assuming global Lipschitz regularity on $F$ and $G$, a mild solution $u$ to (\ref{eqn:SPDE}) exists and is unique \citep[Thm. 6.4]{hairer2009introduction}, at least for short times. 

We follow \citet{friz2020course} and consider a regularization $W^\epsilon = \varphi^\epsilon * W$ of the driving noise $W$ with a  mollifier\footnote{A mollifier $\varphi$ is a smooth function on $\mathbb{R}^{d+1}$ that is: 1) compactly supported, 2) $\int_{\mathbb{R}^{d+1}}\varphi(x)dx = 1$, and 3) $\lim_{\epsilon \to 0} \varphi^\epsilon(x) = \lim_{\epsilon \to 0}\epsilon^{-(d+1)}\varphi(x/\epsilon) = \delta(x)$, where $\delta$ is the Diract delta function and the limit must be understood in the space of Schwartz distributions.}, where $*$ means convolution. As done in \citet{kidger2020neural} for Neural CDEs, we can rewrite the mild solution of the mollified version of \cref{eqn:SPDE} as the following randomly forced PDE
\begin{equation}\label{eqn:SPDE_simple}
    u_t = e^{t\mathcal{L}}u_0 + \int_0^te^{(t-s)\mathcal{L}}H_\xi(u_s)ds, \quad H_\xi(u_t) := F(u_t) + G(u_t)\xi_t,
\end{equation}
where $\xi = \dot{W}^\epsilon$, $\mathcal{H}_u=L^2(\mathcal{D},\mathbb{R}^{d_u})$ and $\mathcal{H}_\xi = L^2(\mathcal{D},\mathbb{R}^{d_\xi})$. We will refer to $\xi$ as \emph{white noise} if $W$ is a cylindrical Wiener process and as \emph{coloured noise} if $W$ is a $Q$-Wiener process.

In view of machine learning applications, one should think of $W$ as a continuous space-time embedding of an underlying spatiotemporal data stream. In this paper we are only going to consider $W$ to be a sample path from a Wiener process, but we emphasise that the Neural SPDE model extends, in principle, beyond the scope of SPDEs and could be used for example to process videos in computer vision applications, which we leave as future work. Next we introduce the Neural SPDE model.

\begin{figure}[h]
    \centering
    \includegraphics[width=0.99\textwidth, trim={0.1cm 0 0.1cm 0}, clip]{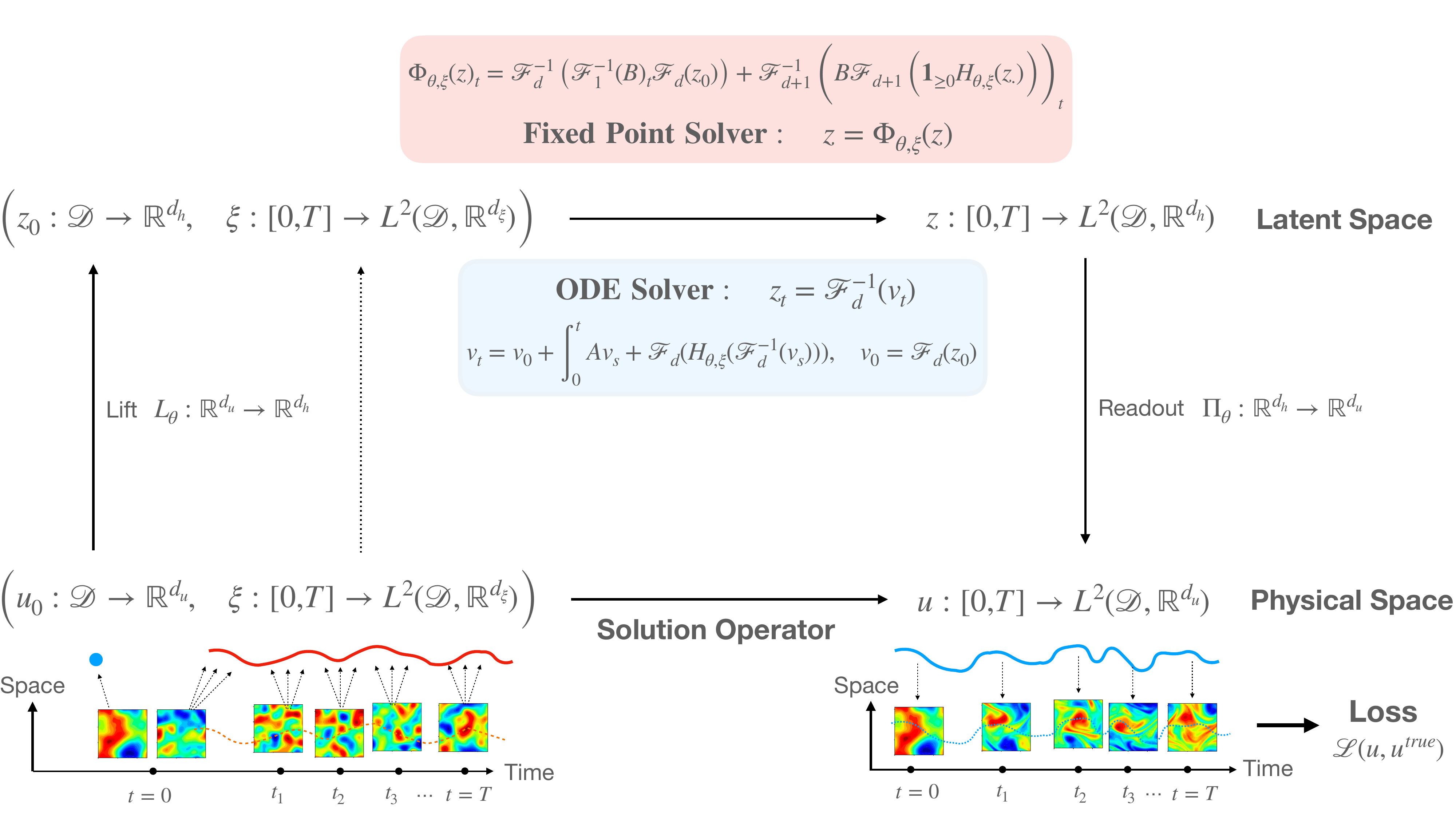}
    \caption{{\small A signal $\xi$ and an initial condition $u_0$ are observed on a, possibly irregular, spatiotemporal grid. $u_0$ is lifted to $z_0$ living in a latent space. The solution of an SPDE in latent space driven by the signal $\xi$ and with initial condition $z_0$ is obtained using either a fixed point solver or an ODE solver. The solution $z$ is then projected back to the physical space by a linear readout. The output signal $u$ is then fed to a pathwise loss function.}}
    \label{fig:spde_scheme}
\end{figure}%

\section{Neural SPDEs}\label{sec:model}

For a large class of differential operators $\mathcal{L}$, the action of the semigroup $e^{t \mathcal{L}}$ can be written as an integral against a kernel function $\mathcal{K}_t:\mathcal{D}\times\mathcal{D} \to \mathbb{R}^{d_u \times d_u}$ such that 
\begin{equation*}
    (e^{t\mathcal{L}}h)(x) = \int_{ \mathcal{D}}\mathcal{K}_t(x,y)h(y)\mu_t(dy),
\end{equation*}
for any $h \in \mathcal{H}_u$, any $x \in \mathcal{D}$ and $t \in [0,T]$, and where $\mu_t$ is a Borel measure on $\mathcal{D}$. As in \citet{kovachki2021neural}, here we take $\mu_t$ to be the Lebesgue measure on $\mathbb{R}^d$ but other choices can be made, for example to incorporate prior information. We assume that $\mathcal{K}$ is stationary so that \cref{eqn:SPDE_simple} can be rewritten in terms of the spatial convolution $*$ 
\begin{equation*}
    u_t = \mathcal{K}_t * u_0 + \int_0^t\mathcal{K}_{t-s} * H_\xi(u_s)ds.
\end{equation*}
For a large class of SPDEs of the form (\ref{eqn:SPDE}), both $F$ and $G$ are local operators acting on a function $h \in \mathcal{H}_u$. In other words, the evaluations $F(h)(x)$ and $G(h)(x)$ at any point $x \in \mathcal{D}$ only depend $h(x)$, and not on the evaluation $h(y)$ at some other point $y \in \mathcal{D}$ in the neighbourhood of $x$. 

\subsection{The model}

Let $\mathcal{H}_h = L^2(\mathcal{D},\mathbb{R}^{d_h})$ for some latent space dimension $d_h>d_u$. Let 
$$L_\theta : \mathbb{R}^{d_u} \to \mathbb{R}^{d_h}, \quad F_\theta : \mathbb{R}^{d_h} \to \mathbb{R}^{d_h}, \quad G_\theta : \mathbb{R}^{d_h} \to \mathbb{R}^{d_h \times d_\xi}, \quad \Pi_\theta : \mathbb{R}^{d_h} \to \mathbb{R}^{d_u}$$ 
be four feedforward neural networks. For any differentiable control $\xi : [0,T] \to \mathcal{H}_\xi$, define the map $H_{\theta,\xi} : \mathcal{H}_u \to \mathcal{H}_u$ so that for any $h \in \mathcal{H}_u$, $x \in \mathcal{D}$, $t \in [0,T]$
\begin{equation*}
    H_{\theta,\xi}(h)(x) = F_\theta(h(x)) + G_\theta(h(x))\xi_t, 
\end{equation*} 
A Neural SPDE is defined as follows
\begin{equation}\label{eqn:NSPDE}
    z_0(x) = L_\theta(u_0(x)), \quad z_t = \mathcal{K}_t * z_0 + \int_0^t\mathcal{K}_{t-s} * H_{\theta,\xi}(z_s)ds, \quad u_t(x) = \Pi_\theta(z_t(x)).
\end{equation}
We note that globally Lipschitz conditions can be imposed by using ReLU or tanh activation functions in the neural networks $F_\theta$ and $G_\theta$. In \cref{ssec:k_1,ssec:k_2} we propose two distinct algorithms to evaluate the Neural SPDE model (\ref{eqn:NSPDE}) which are based on two different parameterization of the kernel $\mathcal{K}$.

\subsection{Evaluating the model by solving a system of ODEs}\label{ssec:k_1}

Levereraging the \emph{convolution theorem}, we can rewrite the integral in \cref{eqn:NSPDE} as follows
\begin{align*}
    z_t = \mathcal{F}^{-1}\Big( & \mathcal{F}(\mathcal{K}_{t}) \mathcal{F}(z_0)  + \int_0^t \mathcal{F}(\mathcal{K}_{t-s})\mathcal{F}(H_{\theta,\xi}(z_s))ds\Big), 
        %\label{eqn:spde_ode}
\end{align*}
where $\mathcal{F}, \mathcal{F}^{-1}$ are the \emph{$d$-dimensional Fourier transform} (FT) and its inverse (see \Cref{def:FT}). If one further assumes that $\mathcal{L}$ is a polynomial differential operator, it can be shown that there exists a map $A :  \mathbb{C}^d \to \mathbb{C}^{d_h \times d_h}$ such that $\mathcal{F}(\mathcal{K}_{t})(y) = e^{tA(y)}$ (see \cref{sec:appendix_ODE_derivation} for a derivation). It follows that 
\begin{equation*}
    z_t = \mathcal{F}^{-1}\Big(e^{tA} \mathcal{F}(z_0) + \int_0^t e^{(t-s)A}\mathcal{F}(H_{\theta,\xi}(z_s))ds\Big) = \mathcal{F}^{-1}(v_t),
\end{equation*}
where $v_t : \mathbb{C}^d \to \mathbb{C}^{d_h}$ is the solution of the following ODE
\begin{align*}
    v_t = v_0 + \int_0^t Av_s + \mathcal{F}(H_{\theta,\xi}(\mathcal{F}^{-1}(v_s))).
\end{align*}
Hence, $z_t$ can be obtained by applying the inverse FT to the output of an ODE solver on $[0,t]$ with initial condition $\mathcal{F}(z_0)$, vector field $\Psi_{\theta,\xi} := A + \mathcal{F} \circ H_{\theta,\xi} \circ \mathcal{F}^{-1}$, i.e.
\begin{equation*}
    z_t \approx \mathcal{F}^{-1}\big(\text{ODESolve}(\mathcal{F}(z_0), \Psi_{\theta,\xi}, [0,t])\big)
\end{equation*}
This approach can naturally be seen as a \say{neural version} of the classical \emph{spectral Galerkin method} for SPDEs as described in \Cref{ssec:num_solvers}. 

We note that this numerical evaluation of the Neural SPDE model (\ref{eqn:NSPDE}) allows to inherit memory-efficient adjoint-based backpropagation capabilities as in the evaluation of a Neural CDE. For further details on adjoint-based backpropagation we refer the reader to \citet{chen2018neural, kidger2022neural}. 

\subsection{Evaluating the model by solving a fixed point problem}\label{ssec:k_2}

In our second approach of model evaluation, we make use of three different versions of the FT: the time-only FT $\mathcal{F}_1$ and its inverse $\mathcal{F}_1^{-1}$, the space-only FT  $\mathcal{F}_d$ and its inverse $\mathcal{F}_d^{-1}$, and the space-time FT $\mathcal{F}_{d+1}$ and its inverse $\mathcal{F}_{d+1}^{-1}$ (see \Cref{def:FT} for details). Denoting by $\star$ the space-time convolution,  the integral in \cref{eqn:NSPDE} can be rewritten as
\begin{equation*}
    z_t = \mathcal{K}_t * z_0 + \left(\mathcal{K} \ \star \ \mathbbm{1}_{\geq 0} H_{\theta,\xi}(z_{\cdot})\right)_t,
\end{equation*}
where $\mathbbm{1}_{\geq 0}$ is the indicator function restricting the temporal domain to the positive real line. Using again the convolution theorem we obtain
\begin{align*}
    z_t &= \mathcal{F}^{-1}_d\left(\mathcal{F}_d(\mathcal{K}_t) \mathcal{F}_d(z_0)\right) + \mathcal{F}^{-1}_{d+1}\left(\mathcal{F}_{d+1}\left(\mathcal{K}\right)\mathcal{F}_{d+1}\left(\mathbbm{1}_{\geq 0}H_{\theta,\xi}(z_{\cdot})\right)\right)_t, \nonumber
\end{align*}
where all multiplications are matrix-vector multiplications. Using the trick introduced in \cite{li2020fourier}, one can parameterize $\mathcal{F}_{d+1}(\mathcal{K})(y)$ directly in Fourier space as a complex tensor $B$, so that the solution of \cref{eqn:NSPDE} can be obtained by solving the fixed point problem $z = \Phi_{\theta, \xi}(z)$ with
\begin{align*}
    \Phi_{\theta, \xi}(z)_t := \mathcal{F}^{-1}_d\left(\mathcal{F}^{-1}_1(B)_t \mathcal{F}_d(z_0)\right) + \mathcal{F}^{-1}_{d+1}\left(B\mathcal{F}_{d+1}\big(\mathbbm{1}_{\geq 0}H_{\theta,\xi}(z_{\cdot})\right)\big)_t, 
\end{align*}
and where we used the fact that $\mathcal{F}_d(\mathcal{K}_t) = \mathcal{F}_1^{-1}(\mathcal{F}_{d+1}(\mathcal{K}))_t$. 
This can be solved numerically using classical root-finding schemes (e.g. by Picard's iteration) 
\begin{equation*}
    z \approx \text{FixedPointSolve}(z_0, \Phi_{\theta,\xi}).
\end{equation*}
Analogously to adjoint-based backpropagation for the evaluation approach mentioned in \Cref{ssec:k_1}, there is a mechanism that leverages the \emph{implicit function theorem} allowing to backpropagate through the operations of a fixed point solver in a memory-efficient way. See \citet{bai2019deep} for further details. 

% In our implementation, we don't use this feature and backpropagate through the operation of the fixed point solver.

% and leaving the implementation of this memory-efficient backpropagation mechanism for training Neural SPDEs as future work.

We note that the FTs are numerically approximated using the \textit{discrete Fourier transform} (DFT) and selecting a maximum number of frequency modes \footnote{The DFT approximates the Fourier series expansion truncated at a maximum number of modes. This allows to specify the shape of the two complex tensors $A$ ($k^1_{\text{max}}\times \ldots\times k^d_{\text{max}}\times d_h\times d_h$) in \Cref{ssec:k_1} and $B$ ($k^1_{\text{max}}\times \ldots\times k^{d+1}_{\text{max}}\times d_h\times d_h$) in \Cref{ssec:k_2}. The $k^i_{\text{max}}$ are treated as hyperparameters of the model.} (see \Cref{ssec:Fourier} for details).

\subsection{Space-time resolution-invariance}

% Let $D = \{x_1,...,x_m\} \subset \mathcal{D}$ be a $m$-points discretization of the spatial domain $\mathcal{D}$, and let $\mathcal{T} = \{t_0, ..., t_n\} \subset [0, T]$ be a $(n+1)$-points discretization of the time interval $[0, T]$ with  $t_0< ...< t_n$. 

As depicted in \Cref{fig:spde_scheme}, the input to a Neural SPDE corresponds to a (possibly irregularly sampled) time-indexed sequence of (possibly partially-observed) spatial observations recorded on a space-time grid. The data is then interpolated into a continuous spatiotemporal signal $\xi$ and initial condition $u_0$. By construction, a Neural SPDE operates in continuous time and space on the tuple of functions $(u_0,\xi)$ and produces a spatiotemporal response $u$, which is also continuous in space and time; the function $u$ can then be evaluated at an arbitrary space-time resolution, possibly different from the one used during training. In the next section we will demonstrate empirically that even if trained on a coarser resolution, a Neural SPDE can be evaluated on a finer resolution without sacrificing performance, a property known as \emph{zero-shot super-resolution}. 

\subsection{Comparison of the two evaluation methods}

\paragraph{Number of parameters} For a fixed dimension $d_h$ of the latent space, the majority of trainable parameters in the ODE parameterization lies in the complex tensor $A$, which consists of $k^1_{\text{max}} ...  k^d_{\text{max}} d_h^2 $ parameters, where each $k^i_{\text{max}}$ is the maximum number of selected frequencies in the Fourier domain. Regarding the Fixed Point parameterization, the bulk of the parameters is in the complex tensor $B$, which consists of $k^1_{\text{max}} ... k^{d+1}_{\text{max}} d_h^2$, where the additional frequency is due to the fact that we are taking the FFT in space-time rather than just in space as done in the ODE Solver approach. Hence, the latter would in principle have the advantage of using a lower number of parameters than the former; however, to achieve similar performance, we found that the dimensionality of the latent space has to be roughly 20 times higher, which offsets the aforementioned advantage.
 
\paragraph{Time complexities} The time complexity of the ODE Solver approach is $\mathcal{O}(N N_x \log(N_x))$, where $N$ is the number of time steps taken by the ODE solver and $N_x$ is the number of points on the spatial grid, while the complexity of the Fixed Point approach is $\mathcal{O}(I N_x N_t (\log(N_x) + \log(N_t)))$ where $I$ is the number of Picard iterations and $N_t$ is the number of points on the temporal grid. In our experiments we choose $N_t \approx N_x$ and $I N_t \approx N$, making the two complexities comparable.
 
\paragraph{Speed of computation} We found that the ODE approach is approximately 10 times slower than the Fixed Point approach. We believe this is largely an implementation issue of the \texttt{torchdiffeq} library, while the FFT is a highly optimised transform in Pytorch.

\subsection{Considerations about convergence}

We follow \citet{friz2020course} and consider a regularization $W^\epsilon = \varphi^\epsilon * W$ of the driving noise $W$ with a compactly supported smooth mollifier $\varphi^\epsilon$. It is a classical result (Wong-Zakai \cite{twardowska1996wong}) from rough path theory \cite{lyons1998differential, gubinelli2004controlling} that, for the case of SDEs, the sequence of random ODEs driven by the mollification of Brownian motion converges in probability to a limiting process that does not depend on the choice of mollifier and agrees with the Stratonovich solution of the SDE. Furthermore, the solution map $(u_0,W) \mapsto u$ is continuous in an appropriate rough path topology. This result nicely extends to the setting of SPDEs driven by a finite dimensional noise \citep[Thm. 1.3]{friz2020course}: if $u^\epsilon$ denotes the random PDE solutions driven by $\dot W^\epsilon dt$ (instead of $\circ dW_t$), then $u^\epsilon$ converges in probability to a limiting process corresponding to the Stratonovich solution of the SPDE. In our setting though, the driving noise is infinite dimensional and the resulting integral cannot be interpreted in the Stratonovich sense because otherwise the corresponding It\^o-Stratonovich correction would be infinite. Nonetheless, \citet[Thm. 1.1]{hairer2015wong} show that, in the case of the heat operator and under appropriate renormalization and drift correction, the random PDE solution $u^\epsilon$ converges in probability to the It\^o solution of the SPDE, and that the solution map is continuous in an appropriate regularity structures topology. We note that extending this result to a generic differential operators would require a similarly rigorous proof, which goes beyond the scope of this article and that we leave as future work.
% Based on our promising empirical results, we think the following are interesting questions for future research: (i) Does the random PDE model in \cref{eqn:SPDE_simple} converge to a limiting It\^o SPDE? Is the solution map continuous in an a regularity structure topology? Is the limiting process independent on the choice of mollifier?

\section{Experiments}\label{sec:experiments}

In this section, we run experiments on three semilinear SPDEs: the stochastic Ginzburg-Landau equation in \ref{ssec:GL}, the stochastic Korteweg-De Vries equation in \ref{ssec:KdV}, and the stochastic Navier-Stokes equations in \ref{ssec:SNS}. We note that although the assumption of globally Lipschitz vector fields might be violated for the following SPDEs, well-posedness (i.e. existence of global solutions) can be shown using equation-specific arguments. We consider three supervised operator-learning settings:
\begin{itemize}
    \item $u_0 \mapsto u$, assuming the noise $\xi$ is not observed;
    \item $\xi \mapsto u$, assuming the noise $\xi$ is observed, but the initial condition $u_0$ is fixed across samples;
    \item $(u_0,\xi) \mapsto u$, assuming the noise $\xi$ is observed and $u_0$ changes across samples.
\end{itemize}

We note that learning the operator $u_0 \mapsto u$ of an SPDE without observing the driving noise $\xi$ unavoidably yields poor results for all considered models as only partial information about the system is provided as input. However, we find it informative to include the performances obtained in this setting, as this provides a sanity check that emphasizes the importance of the noise in all the experiments we consider in this paper. Moreover, the ability to process the initial condition $u_0$ on its own (in absence of noise) testifies that Neural SPDEs can also be used to learn deterministic PDEs. We provide an example on the deterministic Navier-Stokes equations in \Cref{ssec:det_PDEs}.

Neural CDE, Neural RDE, FNO and DeepONet \citep{lu2021learning,lu2022comprehensive} will be the main benchmark models. In addition, we also propose an additional baseline Neural CDE-FNO, which is a hybrid model consisting of a Neural CDE where the drift is modelled by an FNO and the diffusion by a feedforward neural network. The motivation for using a FNO to represent the drift comes from the universal approximation properties of FNOs studied in \citet[Thm. 4]{kovachki2021neural}. 

An interesting line of work to tackle SPDE-learning is provided in \citep{chevyrev2021feature, hu2022neural}. The authors construct a set of features from the
pair $(u_0, \xi)$ following the definition of a model from the theory of regularity structures \citep{hairer2014theory}. They then perform linear \citep{chevyrev2021feature} and nonlinear \citep{hu2022neural} regression from these features to the
solution of the SPDE at a single time point. Therefore, these models would have to be retrained for any new prediction. In addition, both \citep{chevyrev2021feature, hu2022neural} assume knowledge of the differential operator $\mathcal{L}$ governing the dynamics, while Neural SPDE learns a representation of $\mathcal{L}$ via the parametrization of the associated kernel. For these reasons these recent models are not included in our benchmark.

For all the experiments, the loss function is the relative pathwise $L^2$ error. The hyper-parameters for all the models are selected by grid-search (see \Cref{ssec:experimental_details} for further experimental details). Experiments are run on a Tesla P100 NVIDIA GPU. The code for the experiments is provided in the supplementary material. Additional experiments may be found in \Cref{sec:experiments_appendix}.

% In the Ginzburg-Landau example, for all models we use $d_h=16$ latent units. For FNO and Neural SPDE we use the $32$ maximal frequency modes both in space and time. For the Navier-Stokes example, we use the maximal $16$ frequencies in space, the maximal $10$ frequencies in time and $d_h=8$. 

% \subsection{Complexity analysis}
% The time complexity of the fast Fourier transform (FFT) of a signal is $\mathcal{O}(L\log L)$ where $L$ is the number of points used to compute the FFT. The computational bottleneck of Neural SPDEs is the computation of the space-time fast Fourier transform of a space-time signal evolving in a $d_h$-dimensional space. The time complexity is $\mathcal{O}(d_hL\log L)$ where $L=\prod_{i=1}^{d}L_i$ with $d$ the number of space-time dimensions ($d=2$ or $d=3$ in our experiments).  

% \begin{figure*}[h]
%     \centering
%     \includegraphics[scale=0.27]{fig/Phi41.pdf}
%     \caption{Ground truth and predictions on three test instances for the Ginzburg-Landau example.}
%     \label{fig:Phi41}
% \end{figure*}

\subsection{Stochastic Ginzburg-Landau equation}\label{ssec:GL}

We start with the stochastic Ginzburg-Landau equation, a reaction diffusion equation in 1D given by
\begin{align*}
    \partial_t u - \Delta u = 3u -u^3 + \xi,  \qquad u(t,0) = u(t,1), \quad
    u(0,x) = u_0(x), \quad (t,x)\in [0,T]\times[0,1]. 
\end{align*}
This equation is also known as the Allen-Cahn equation in $1$-dimension and is used for modeling various physical phenomena like superconductivity \citep{temam2012infinite}. Here $\xi$ denotes space-time white noise with sample paths generated using classical sampling schemes for Wiener processes detailed in \ref{ssec:wiener}. 

\begin{table*}[h]
\caption{{\small\textbf{Ginzburg-Landau}. Relative L2 error on the test set. x indicates that the model is not applicable. }}
%\textit{In the first task ($u_0\mapsto u$) only partial information ($u_0$) is provided as input; the underlying noise $\xi$ is used to generate the dynamics, it varies across samples, but is not provided as an input to the models.} This explains why the performance of the applicable models is poorer than for all other tasks. In the second task ($\xi\mapsto u$) the initial condition $u_0$ is kept fixed. In the third task, the initial condition $u_0$ and the noise $\xi$ change and are both provided as input.} 
\begin{center}
\begin{tabular}{lccc|ccc}
\toprule
   \multirowcell{2}[0pt][l]{Model} &  \multicolumn{3}{c}{$N=1\,000$} &  \multicolumn{3}{c}{$N=10\,000$} \\
    \cmidrule(lr){2-7} 
 & $u_0\mapsto u$ & $\xi\mapsto u$ & $(u_0,\xi)\mapsto u$ & $u_0\mapsto u$ & $\xi\mapsto u$ & $(u_0,\xi)\mapsto u$\\
\midrule 
% NODE &  & x & x & & x & x \\
NCDE & x & 0.112 & 0.127 & x & 0.056 & 0.072 \\
NRDE & x & 0.129 & 0.150 & x & 0.070 & 0.083 \\
% NCDE-spline & x & 0.110 & 0.128 & x &  0.056 & 0.070\\
NCDE-FNO & x & 0.071 & 0.066 & x & 0.066 &  0.069 \\
% NRDE (depth 2) & x & x & x & x & x & x & x & x\\
DeepONet & 0.130 & 0.126 & x & 0.126 & 0.061 & x \\
FNO & 0.128 & 0.032 & x & 0.126 & 0.027 & x \\
% \textcolor{red}{another FNO} & & & & & & & \\
\midrule
NSPDE (Ours)  &  0.128 & \textbf{0.009} & \textbf{0.012} & 0.126 & \textbf{0.006} &  \textbf{0.006} \\
\bottomrule
\end{tabular}
\label{table:phi41}
\end{center}
\end{table*}

We consider two data-regimes: a \emph{low data regime} where the total number of training observations is $N=1\,000$, and a \emph{large data regime} where $N=10\,000$. In both cases, the response paths are generated by solving the SPDE along each sample path of the noise $\xi$ using a finite difference scheme described in \Cref{ssec:num_solvers} using $128$ evenly distanced points in space and time and step size $\Delta t=10^{-3}$. Following the same setup as in \citet[eq. (3.6)]{chevyrev2021feature}, we solve the SPDE until $T=0.05$ resulting in $50$ time points . We choose as initial condition $u_0(x)=x(1-x)+\kappa\eta(x)$, with $\eta(x)=a_0 + \sum_{k=-10}^{k=10}a_k/(1+|k|^2)\sin\left(k\pi x\right)$ where $a_k\sim\mathcal{N}(0,1)$. We take $\kappa=0$ and $\kappa=0.1$ to generate a dataset where the initial data is either fixed or varies across samples. We provide extra experiments on this SPDE for larger time horizons $T$ and multiplicative forcing in \Cref{ssec:GLE}. We report the results in \Cref{table:phi41}. The Neural SPDE model (NSPDE) yields the lowest relative error for all tasks, reaching one order of magnitude improvement on the main task $(u_0,\xi) \mapsto u$ in the large data regime compared to all the applicable benchmark models (NCDE, NRDE, NCDE-FNO). In all settings, even with a limited amount of training samples ($N=1\,000$), NSPDE achieves $\sim1\%$ error rate, and marginally improves to $<1\%$ error when $N=10\,000$.

\subsection{Stochastic Korteweg–De Vries equation}\label{ssec:KdV}

Next, we consider the stochastic Korteweg–De Vries (KdV) equation, a higher order SPDE given by
\begin{align*}
    \partial_t u + \gamma\partial_x^3 u = 6u\partial_xu + \xi, \qquad
    u(t,0) = u(t,1), \quad
    u(0,x) = u_0(x), \quad (t,x)\in [0,T]\times[0,1]\,.
\end{align*}
This equation is used to describe the propagation of nonlinear waves at the surface of a fluid subject to random perturbations (another wave equation is studied in~\Cref{ssec:Wave}). We refer the reader to~\citet{wazwaz2009solitary} for an overview on the KdV equation and its relations to solitary waves. The stochastic forcing is given by $\xi = \dot{W}$ for $W$ being a partial sum approximation of a Q-Wiener process as per Example 10.8 in~\citet{lord2014introduction} with $\lambda_j \sim j^{-5+\varepsilon}$ and $\phi_j(x) = \sin{(j\pi x)}$ (see \cref{eqn:expansion_q_wiener} in \Cref{ssec:wiener}). Taking small $\varepsilon > 0$ guarantees that $W_t$ is twice differentiable in space for every $t \geq 0$. To generate the datasets, we solve the SPDE with $\gamma=0.1$ until $T=0.5$. 

\begin{wraptable}[28]{l}{0.55\textwidth}
\vspace{-1.\baselineskip}
\caption{{\small\textbf{Stochastic KdV}. Relative L2 error on the test set. The symbol x indicates that the model is not applicable.}}
% \begin{subtable}[h]{0.45\textwidth}
\begin{subtable}[h]{.5\textwidth}
\caption{$N=1\,000$ and $T=0.5$.}
\begin{tabular}{lccc}
\toprule
  Model &  $u_0\mapsto u$ & $\xi\mapsto u$ & $(u_0,\xi)\mapsto u$\\
\midrule 
NCDE & x & 0.464 & 0.466  \\
NRDE & x & 0.497 & 0.503  \\
NCDE-FNO & x & 0.126 & 0.259  \\
DeepONet & 0.874 & 0.235 & x \\
FNO & 0.835 & 0.079 & x  \\
\midrule
NSPDE (Ours) & 0.832 &  \textbf{0.004} & \textbf{0.008} \\
\bottomrule
\end{tabular}
\label{table:kdv}%
\end{subtable}
\newline
\vspace{1\baselineskip}
\newline
\begin{subtable}[h]{.5\textwidth}
\caption{$N=1\,000$ and $T=1$.}
\begin{tabular}{lccc}
\toprule
  Model & $u_0\mapsto u$ & $\xi\mapsto u$ & $(u_0,\xi)\mapsto u$ \\
\midrule 
FNO &  0.913 &  0.112 & x  \\
\midrule
NSPDE (Ours) & 0.904 & \textbf{0.009} & 0.012   \\
\bottomrule
\end{tabular}
\label{table:kdvfull}
\end{subtable}
\newline
\vspace{1\baselineskip}
\newline
\begin{subtable}[h]{.5\textwidth}
\caption{Subsampling ($N=1\,000$ and $T=0.5$).}
\begin{tabular}{llccc}
\toprule
\multicolumn{2}{c}{Subsampling rates} & $\xi\mapsto u$ & $(u_0,\xi)\mapsto u$ \\
\midrule
 $\text{Time}:0\%$ & $\text{Space}:0\%$ & 0.004 &   0.008  \\
 $\text{Time}:10\%$ & $\text{Space}:0\%$ &  0.076 & 0.059  \\
 $\text{Time}:0\%$ & $\text{Space}:50\%$ & 0.005 &  0.008\\
\bottomrule
\end{tabular}
\label{table:kdv_subsampling}
\end{subtable}
\end{wraptable}
The stochastic forcing is simulated using $128$ evenly distanced points in space and a time step $\Delta t_{\text{ref}}=10^{-3}$. We then approximate realizations of the solution of the KdV equation using a time step $\Delta t=10^{-2}$ until $T=0.5$. Here, the initial condition is given by $u_0(x)=\sin(2\pi x)+\kappa\eta(x)$, where $\eta$ is defined as in \Cref{ssec:GL}. Similarly to \Cref{ssec:GL} we either take $\kappa=0$ or $\kappa=1$ to generate datasets where the initial condition is either fixed or varies across samples. Each dataset consists of $N=1\,000$ training observations. As reported in \Cref{table:kdv}, Neural SPDEs outperforms the second best model FNO by a full order of magnitude in the task $\xi \mapsto u$ and the second best model NCDE-FNO by almost two orders of magnitude in the task $(u_0,\xi) \mapsto u$. We also perform the same tasks for a larger time horizon $T=1$ and report the results of a comparison against FNO in \Cref{table:kdvfull}.

\paragraph{Partial observations} 
Neural SPDEs are able to process signals that are irregularly sampled both in space and in time by interpolating between observations. Yet, the ability of a model to process irregular data does not guarantee its robustness when some observations are dropped. Robustness can only be guaranteed if the  signal is regular enough so that replacing dropped observations by interpolation results in a new signal that is close, in some suitable norm, to the original signal. To illustrate this point, we run two additional experiments where we drop uniformly at random 1) 10\% of the data in time and 2) 50\% of the data in space. As it can be observed in \Cref{table:kdv_subsampling}, the performance of Neural SPDE remains roughly unchanged when data is dropped in space but decreases when data is dropped in time, which is to be expected since the driving signal is a Q-Wiener process, which is rough in time, but smoother in space. We also note that to ensure a good approximation of the FT by the FFT, the interpolation must translate the irregular data to a (possibly finer) regular grid.

\subsection{Stochastic Navier-Stokes equations  in 2D}\label{ssec:SNS}
Finally, we consider the vorticity form of the Navier-Stokes equations for an incompressible flow
\begin{align}\label{eq:SNS}
    \partial_t w- \nu \Delta w= - u\cdot \nabla w + f +\sigma\xi, \qquad
    w(0,x) = w_0(x), \quad (t,x) \in [0,T] \times [0,1]^2, 
\end{align} 
where $u$ is the unique divergence free ($\nabla \cdot u = 0$) velocity field such that $w=\nabla\times u$. These equations describe the motion of an incompressible fluid with viscosity $\nu$ subject to external forces \citep{temam2012infinite}. The deterministic forcing $f$, defined as in \citet{li2020fourier}, is a function of space only. The stochastic forcing $\xi$ is given by $\xi = \dot{W}$ for $W$ being a Q-Wiener process which is colored in space and rescaled by $\sigma=0.05$ (see \Cref{ssec:wiener}).
The initial condition is generated according to $w_0\sim\mathcal{N}(0, 3^{3/2}(-\Delta+49I)^{-3})$
with periodic boundary conditions. The viscosity  is set to $\nu=10^{-4}$. 

\begin{figure*}[h]
    \centering
    \includegraphics[scale=0.35]{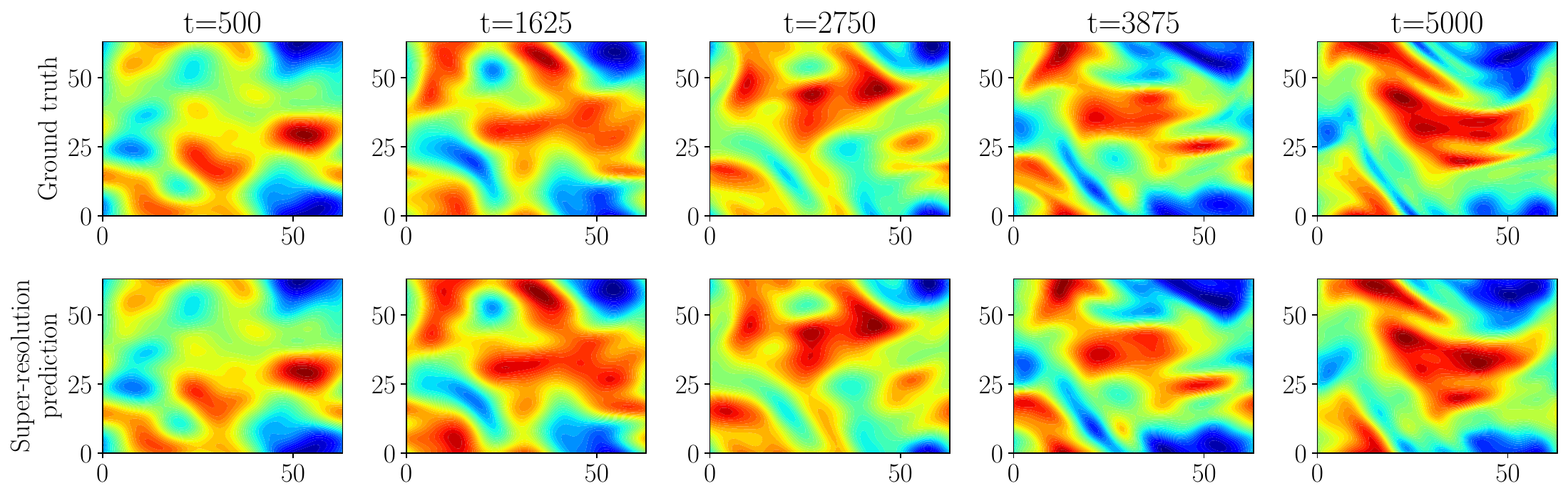}
    \caption{{\small\textbf{Top panel:} Solution of the vorticity equation for one realisation of the stochastic forcing between the $500^{\text{th}}$ and the $5\,000^{\text{th}}$ time steps.  \textbf{Bottom panel:} Predictions with the Neural SPDE model given the initial condition at the $500^{\text{th}}$ time step and the forcing between the $500^{\text{th}}$ and the $5\,000^{\text{th}}$ time steps. The model is trained on a $16\times 16$ mesh and evaluated on a $64\times 64$ mesh.}}
    \label{fig:mySNS}
\end{figure*}
For each realization of the Q-Wiener process (sampled according to the scheme in \Cref{ssec:wiener}) we solve \cref{eq:SNS} with a pseudo-spectral solver described in \Cref{ssec:num_solvers}, where time is advanced with a Crank–Nicolson update. We solve the SPDE on a $64\times64$ mesh in space and use a time step of size $10^{-3}$. For the tasks $u_0\mapsto u $ and $\xi\mapsto u$, we generate the datasets by solving the SPDE up to time $T=1$ and downsample the trajectories by a factor of $10$ in time (resulting in $100$ time steps) and $4$ in space (resulting in a $16\times16$ spatial resolution). The number of training samples is $N=1\,000$. To generate the training set for the task $(u_0, \xi)\mapsto u$, we generate $10$ long trajectories of $15\,000$ steps each up to time $T=15$. We partition each trajectory into consecutive sub-trajectories of $500$ time-steps using a rolling window. This yields a total of $2\,000$ input-output pairs. We split the data into shorter sequences of $500$ time steps so that one batch fits in memory of the used GPU. 
\begin{wraptable}[15]{l}{0.55\textwidth}
\vspace{0.4\baselineskip}
\caption{{\small\textbf{Stochastic Navier-Stokes}. Relative L2 error on the test set. The symbol x indicates that the model is not applicable, and - indicates that the model does not fit in memory.}}
\begin{tabular}{lccc}
\toprule
  Model &  $u_0\mapsto u$ & $\xi\mapsto u$ & $(u_0,\xi)\mapsto u$ \\
\midrule 
NCDE & x & 0.366 & 0.843 \\
NRDE & x & - & - \\
NCDE-FNO & x & 0.326 & 0.178 \\
% NRDE (depth 2) & x & x & x &  x\\
DeepONet & 0.432  & 0.348 & x \\
FNO & 0.188 & 0.039 & x \\
\midrule
% NSPDE ($n=1$) (Ours)  &   &  0.043 &   &  $\pm$ \\
NSPDE (Ours)  & 0.155  & \textbf{0.034} & \textbf{0.049} \\
\bottomrule
\end{tabular}
\label{table:sns}
\end{wraptable}

As shown in \Cref{table:sns}, Neural SPDEs marginally outperforms FNO on the task $\xi \mapsto u$, but with a significantly larger gap on the task $(u_0,\xi) \mapsto u$ from the second best model NCDE-FNO. \Cref{fig:mySNS} indicates that our model is capable of zero-shot super-resolution in space-time, achieving good performance even when evaluated on a larger time horizon and on an upsampled spatial grid. Finally, we report in \Cref{table:runtime} some run time statistics indicating that NSPDEs can be up to 3 orders of magnitude faster than traditional numerical solvers.
\begin{wraptable}[12]{r}{0.35\textwidth}
% \begin{table}[h]
\vspace{-1.\baselineskip}
\caption{{\small Ratio of the inference time of the trained NSPDE over the runtime of the numerical solver. We use the same spatiotemporal discretization for both NSPDE and the numerical solver. 
% For the Ginzburg-Landau equation and the stochastic wave equation we use a grid with $1\,000$ time points in $[0,1]$ and $128$ space points in $[0,1]$. For the stochastic Navier-Stokes equation we use $1\,000$ time points in $[0,1]$ and $64\times 64$ space points in $[0,1]\times[0,1]$.
}}
% \begin{center}
\begin{tabular}{lc}
\toprule
Dataset & Speedup \\ 
\midrule 
Ginzburg-Landau & 59$\times$\\
% Stochastic Wave Equation & 88$\times$\\
Korteweg-De Vries & 80$\times$\\
Navier-Stokes & 300$\times$\\
\bottomrule
\end{tabular}
\label{table:runtime}
% \end{center}
\end{wraptable}
% \end{table}

\section{Conclusion}\label{sec:conclusion}

We introduced Neural SPDEs, a model capable of learning solution operators of PDEs with (possibly stochastic) forcing from partially observed data. Our model provides an extension to two classes of physics-inspired models. It extends Neural CDEs in that it is resolution-invariant both in space and in time, and it extends Neural Operators as it can be used to learn solution operators of SPDEs depending simultaneously on the initial condition and driving noise. We performed extensive experiments illustrating how the model achieves superior performance while requiring a lower amount of training data compared to other models, and its evaluation is up to 3 orders of magnitude faster than traditional numerical solvers.

\paragraph{Limitations and future work} Similarly to other neural operator models, parameterising the kernel in Fourier space is by no means the only available option; other parameterisations could mitigate some of the disadvantages of the FFT  (irregular grids, aliasing effect ...), see for example \cite{li2020neural}. A question we leave to future work is how to construct a discrepancy between probability measures supported on spatiotemporal signals, generalizing for example the signature kernel MMD in \cite{salvi2021signature}. Neural SPDEs paired with such discrepancy would allow the design of new generative models for spatiotemporal signals. Another research direction will be to assess whether Neural SPDEs can be used in computer vision to process videos at arbitrary resolution.

\begin{ack}
This project was supported by G-Research and by DataSig under the grant EP/S026347/1.
\end{ack}

% \section*{References}
% \newpage

\bibliographystyle{plainnat}
\bibliography{references}

\begin{thebibliography}{38}
\providecommand{\natexlab}[1]{#1}
\providecommand{\url}[1]{\texttt{#1}}
\expandafter\ifx\csname urlstyle\endcsname\relax
  \providecommand{\doi}[1]{doi: #1}\else
  \providecommand{\doi}{doi: \begingroup \urlstyle{rm}\Url}\fi

\bibitem[Alimov et~al.(1992)Alimov, Ashurov, and Pulatov]{alimov1992multiple}
Sh~A Alimov, RR~Ashurov, and AK~Pulatov.
\newblock Multiple fourier series and fourier integrals.
\newblock In \emph{Commutative Harmonic Analysis IV}, pages 1--95. Springer,
  1992.

\bibitem[Bai et~al.(2019)Bai, Kolter, and Koltun]{bai2019deep}
Shaojie Bai, J~Zico Kolter, and Vladlen Koltun.
\newblock Deep equilibrium models.
\newblock \emph{Advances in Neural Information Processing Systems},
  32:\penalty0 690--701, 2019.

\bibitem[Bellot and Van Der~Schaar(2021)]{bellot2021policy}
Alexis Bellot and Mihaela Van Der~Schaar.
\newblock Policy analysis using synthetic controls in continuous-time.
\newblock In \emph{International Conference on Machine Learning}, pages
  759--768. PMLR, 2021.

\bibitem[Briggs and Henson(1995)]{briggs1995dft}
William~L Briggs and Van~Emden Henson.
\newblock \emph{The DFT: an owner's manual for the discrete Fourier transform}.
\newblock SIAM, 1995.

\bibitem[Chen et~al.(2018)Chen, Rubanova, Bettencourt, and
  Duvenaud]{chen2018neural}
Ricky~TQ Chen, Yulia Rubanova, Jesse Bettencourt, and David Duvenaud.
\newblock Neural ordinary differential equations.
\newblock In \emph{Proceedings of the 32nd International Conference on Neural
  Information Processing Systems}, pages 6572--6583, 2018.

\bibitem[Chen and Chen(1995)]{chen1995universal}
Tianping Chen and Hong Chen.
\newblock Universal approximation to nonlinear operators by neural networks
  with arbitrary activation functions and its application to dynamical systems.
\newblock \emph{IEEE Transactions on Neural Networks}, 6\penalty0 (4):\penalty0
  911--917, 1995.

\bibitem[Chevyrev et~al.(2021)Chevyrev, Gerasimovics, and
  Weber]{chevyrev2021feature}
Ilya Chevyrev, Andris Gerasimovics, and Hendrik Weber.
\newblock Feature engineering with regularity structures.
\newblock \emph{arXiv preprint arXiv:2108.05879}, 2021.

\bibitem[Cooley and Tukey(1965)]{cooley1965algorithm}
James~W Cooley and John~W Tukey.
\newblock An algorithm for the machine calculation of complex fourier series.
\newblock \emph{Mathematics of computation}, 19\penalty0 (90):\penalty0
  297--301, 1965.

\bibitem[Friz and Hairer(2020)]{friz2020course}
Peter~K Friz and Martin Hairer.
\newblock \emph{A course on rough paths}.
\newblock Springer, 2020.

\bibitem[Gubinelli(2004)]{gubinelli2004controlling}
Massimiliano Gubinelli.
\newblock Controlling rough paths.
\newblock \emph{Journal of Functional Analysis}, 216\penalty0 (1):\penalty0
  86--140, 2004.

\bibitem[Hairer(2009)]{hairer2009introduction}
Martin Hairer.
\newblock An introduction to stochastic pdes.
\newblock \emph{arXiv preprint arXiv:0907.4178}, 2009.

\bibitem[Hairer(2013)]{hairer2013solving}
Martin Hairer.
\newblock Solving the kpz equation.
\newblock \emph{Annals of mathematics}, pages 559--664, 2013.

\bibitem[Hairer(2014)]{hairer2014theory}
Martin Hairer.
\newblock A theory of regularity structures.
\newblock \emph{Inventiones mathematicae}, 198\penalty0 (2):\penalty0 269--504,
  2014.

\bibitem[Hairer and Pardoux(2015)]{hairer2015wong}
Martin Hairer and {\'E}tienne Pardoux.
\newblock A wong-zakai theorem for stochastic pdes.
\newblock \emph{Journal of the Mathematical Society of Japan}, 67\penalty0
  (4):\penalty0 1551--1604, 2015.

\bibitem[Holden et~al.(1996)Holden, {\O}ksendal, Ub{\o}e, and
  Zhang]{holden1996stochastic}
Helge Holden, Bernt {\O}ksendal, Jan Ub{\o}e, and Tusheng Zhang.
\newblock Stochastic partial differential equations.
\newblock In \emph{Stochastic partial differential equations}, pages 141--191.
  Springer, 1996.

\bibitem[Hu et~al.(2022)Hu, Meng, Chen, Gong, Wang, Chen, Zhu, Ma, and
  Liu]{hu2022neural}
Peiyan Hu, Qi~Meng, Bingguang Chen, Shiqi Gong, Yue Wang, Wei Chen, Rongchan
  Zhu, Zhi-Ming Ma, and Tie-Yan Liu.
\newblock Neural operator with regularity structure for modeling dynamics
  driven by spdes.
\newblock \emph{arXiv preprint arXiv:2204.06255}, 2022.

\bibitem[Kidger(2022)]{kidger2022neural}
Patrick Kidger.
\newblock On neural differential equations.
\newblock \emph{arXiv preprint arXiv:2202.02435}, 2022.

\bibitem[Kidger et~al.(2020)Kidger, Morrill, Foster, and
  Lyons]{kidger2020neural}
Patrick Kidger, James Morrill, James Foster, and Terry Lyons.
\newblock Neural controlled differential equations for irregular time series.
\newblock \emph{arXiv preprint arXiv:2005.08926}, 2020.

\bibitem[Kidger et~al.(2021{\natexlab{a}})Kidger, Foster, Li, and
  Lyons]{kidger2021efficient}
Patrick Kidger, James Foster, Xuechen Li, and Terry Lyons.
\newblock Efficient and accurate gradients for neural sdes.
\newblock \emph{arXiv preprint arXiv:2105.13493}, 2021{\natexlab{a}}.

\bibitem[Kidger et~al.(2021{\natexlab{b}})Kidger, Foster, Li, Oberhauser, and
  Lyons]{kidger2021neural}
Patrick Kidger, James Foster, Xuechen Li, Harald Oberhauser, and Terry Lyons.
\newblock Neural sdes as infinite-dimensional gans.
\newblock \emph{arXiv preprint arXiv:2102.03657}, 2021{\natexlab{b}}.

\bibitem[Kovachki et~al.(2021)Kovachki, Li, Liu, Azizzadenesheli, Bhattacharya,
  Stuart, and Anandkumar]{kovachki2021neural}
Nikola Kovachki, Zongyi Li, Burigede Liu, Kamyar Azizzadenesheli, Kaushik
  Bhattacharya, Andrew Stuart, and Anima Anandkumar.
\newblock Neural operator: Learning maps between function spaces.
\newblock \emph{arXiv preprint arXiv:2108.08481}, 2021.

\bibitem[Li et~al.(2020{\natexlab{a}})Li, Wong, Chen, and
  Duvenaud]{li2020scalable}
Xuechen Li, Ting-Kam~Leonard Wong, Ricky~TQ Chen, and David Duvenaud.
\newblock Scalable gradients for stochastic differential equations.
\newblock In \emph{International Conference on Artificial Intelligence and
  Statistics}, pages 3870--3882. PMLR, 2020{\natexlab{a}}.

\bibitem[Li et~al.(2020{\natexlab{b}})Li, Kovachki, Azizzadenesheli, Liu,
  Bhattacharya, Stuart, and Anandkumar]{li2020neural}
Zongyi Li, Nikola Kovachki, Kamyar Azizzadenesheli, Burigede Liu, Kaushik
  Bhattacharya, Andrew Stuart, and Anima Anandkumar.
\newblock Neural operator: Graph kernel network for partial differential
  equations.
\newblock \emph{arXiv preprint arXiv:2003.03485}, 2020{\natexlab{b}}.

\bibitem[Li et~al.(2020{\natexlab{c}})Li, Kovachki, Azizzadenesheli, Liu,
  Stuart, Bhattacharya, and Anandkumar]{li2020multipole}
Zongyi Li, Nikola Kovachki, Kamyar Azizzadenesheli, Burigede Liu, Andrew
  Stuart, Kaushik Bhattacharya, and Anima Anandkumar.
\newblock Multipole graph neural operator for parametric partial differential
  equations.
\newblock \emph{Advances in Neural Information Processing Systems}, 33,
  2020{\natexlab{c}}.

\bibitem[Li et~al.(2020{\natexlab{d}})Li, Kovachki, Azizzadenesheli,
  Bhattacharya, Stuart, Anandkumar, et~al.]{li2020fourier}
Zongyi Li, Nikola~Borislavov Kovachki, Kamyar Azizzadenesheli, Kaushik
  Bhattacharya, Andrew Stuart, Anima Anandkumar, et~al.
\newblock Fourier neural operator for parametric partial differential
  equations.
\newblock In \emph{International Conference on Learning Representations},
  2020{\natexlab{d}}.

\bibitem[Liu et~al.(2019)Liu, Xiao, Si, Cao, Kumar, and Hsieh]{liu2019neural}
Xuanqing Liu, Tesi Xiao, Si~Si, Qin Cao, Sanjiv Kumar, and Cho-Jui Hsieh.
\newblock Neural sde: Stabilizing neural ode networks with stochastic noise.
\newblock \emph{arXiv preprint arXiv:1906.02355}, 2019.

\bibitem[Lord et~al.(2014)Lord, Powell, and Shardlow]{lord2014introduction}
Gabriel~J Lord, Catherine~E Powell, and Tony Shardlow.
\newblock \emph{An introduction to computational stochastic PDEs}, volume~50.
\newblock Cambridge University Press, 2014.

\bibitem[Lu et~al.(2021)Lu, Jin, Pang, Zhang, and Karniadakis]{lu2021learning}
Lu~Lu, Pengzhan Jin, Guofei Pang, Zhongqiang Zhang, and George~Em Karniadakis.
\newblock Learning nonlinear operators via deeponet based on the universal
  approximation theorem of operators.
\newblock \emph{Nature Machine Intelligence}, 3\penalty0 (3):\penalty0
  218--229, 2021.

\bibitem[Lu et~al.(2022)Lu, Meng, Cai, Mao, Goswami, Zhang, and
  Karniadakis]{lu2022comprehensive}
Lu~Lu, Xuhui Meng, Shengze Cai, Zhiping Mao, Somdatta Goswami, Zhongqiang
  Zhang, and George~Em Karniadakis.
\newblock A comprehensive and fair comparison of two neural operators (with
  practical extensions) based on fair data.
\newblock \emph{Computer Methods in Applied Mechanics and Engineering},
  393:\penalty0 114778, 2022.

\bibitem[Lyons(1998)]{lyons1998differential}
Terry~J Lyons.
\newblock Differential equations driven by rough signals.
\newblock \emph{Revista Matem{\'a}tica Iberoamericana}, 14\penalty0
  (2):\penalty0 215--310, 1998.

\bibitem[Mikulevicius and Rozovskii(2004)]{mikulevicius2004stochastic}
Remigijus Mikulevicius and Boris~L Rozovskii.
\newblock Stochastic navier--stokes equations for turbulent flows.
\newblock \emph{SIAM Journal on Mathematical Analysis}, 35\penalty0
  (5):\penalty0 1250--1310, 2004.

\bibitem[Morrill et~al.(2021)Morrill, Salvi, Kidger, and
  Foster]{morrill2021neural}
James Morrill, Cristopher Salvi, Patrick Kidger, and James Foster.
\newblock Neural rough differential equations for long time series.
\newblock In \emph{International Conference on Machine Learning}, pages
  7829--7838. PMLR, 2021.

\bibitem[Salvi et~al.(2021)Salvi, Cass, Foster, Lyons, and
  Yang]{salvi2021signature}
Cristopher Salvi, Thomas Cass, James Foster, Terry Lyons, and Weixin Yang.
\newblock The signature kernel is the solution of a goursat pde.
\newblock \emph{SIAM Journal on Mathematics of Data Science}, 3\penalty0
  (3):\penalty0 873--899, 2021.

\bibitem[Temam(2012)]{temam2012infinite}
Roger Temam.
\newblock \emph{Infinite-dimensional dynamical systems in mechanics and
  physics}, volume~68.
\newblock Springer Science \& Business Media, 2012.

\bibitem[Twardowska(1996)]{twardowska1996wong}
Krystyna Twardowska.
\newblock Wong-zakai approximations for stochastic differential equations.
\newblock \emph{Acta Applicandae Mathematica}, 43\penalty0 (3):\penalty0
  317--359, 1996.

\bibitem[Wazwaz(2009)]{wazwaz2009solitary}
Abdul-Majid Wazwaz.
\newblock Solitary waves theory.
\newblock In \emph{Partial Differential Equations and Solitary Waves Theory},
  pages 479--502. Springer, 2009.

\bibitem[Weinan(2017)]{weinan2017proposal}
E~Weinan.
\newblock A proposal on machine learning via dynamical systems.
\newblock \emph{Communications in Mathematics and Statistics}, 1\penalty0
  (5):\penalty0 1--11, 2017.

\bibitem[Young(1905)]{young1905vi}
William~Henry Young.
\newblock Vi. on the general theory integration.
\newblock \emph{Philosophical Transactions of the Royal Society of London.
  Series A, Containing Papers of a Mathematical or Physical Character},
  204\penalty0 (372-386):\penalty0 221--252, 1905.

\end{thebibliography}

%%%%%%%%%%%%%%%%%%%%%%%%%%%%%%%%%%%%%%%%%%%%%%%%%%%%%%%%%%%%
% \section*{Checklist}

% %%% BEGIN INSTRUCTIONS %%%
% The checklist follows the references.  Please
% read the checklist guidelines carefully for information on how to answer these
% questions.  For each question, change the default \answerTODO{} to \answerYes{},
% \answerNo{}, or \answerNA{}.  You are strongly encouraged to include a {\bf
% justification to your answer}, either by referencing the appropriate section of
% your paper or providing a brief inline description.  For example:
% \begin{itemize}
%   \item Did you include the license to the code and datasets? \answerYes{See Section 1.}
%   \item Did you include the license to the code and datasets? \answerNo{The code and the data are proprietary.}
%   \item Did you include the license to the code and datasets? \answerNA{}
% \end{itemize}
% Please do not modify the questions and only use the provided macros for your
% answers.  Note that the Checklist section does not count towards the page
% limit.  In your paper, please delete this instructions block and only keep the
% Checklist section heading above along with the questions/answers below.
% %%% END INSTRUCTIONS %%%

\begin{checklist}
\begin{enumerate}

\item For all authors...
\begin{enumerate}
  \item Do the main claims made in the abstract and introduction accurately reflect the paper's contributions and scope?
    \answerYes{See the paragraph entitled "Contributions" in \Cref{sec:intro} for a clear statement of the contributions, and \Cref{sec:background} for further details on the paper's scope.}
  \item Did you describe the limitations of your work?
    \answerYes{See \Cref{sec:background} and \Cref{sec:conclusion}.}
  \item Did you discuss any potential negative societal impacts of your work?
    \answerNo{}
  \item Have you read the ethics review guidelines and ensured that your paper conforms to them?
    \answerYes{}
\end{enumerate}

\item If you are including theoretical results...
\begin{enumerate}
  \item Did you state the full set of assumptions of all theoretical results?
    \answerYes{See \Cref{sec:background}.}
        \item Did you include complete proofs of all theoretical results? \answerNA{}

\end{enumerate}

\item If you ran experiments...
\begin{enumerate}
  \item Did you include the code, data, and instructions needed to reproduce the main experimental results (either in the supplemental material or as a URL)?
    \answerYes{}
  \item Did you specify all the training details (e.g., data splits, hyperparameters, how they were chosen)?
    \answerYes{See \Cref{ssec:experimental_details}.}
        \item Did you report error bars (e.g., with respect to the random seed after running experiments multiple times)?
    \answerNo{Error bars are not reported because it would be too computationally expensive for the baseline models NCDE, NRDE and NCDE-FNO.}
        \item Did you include the total amount of compute and the type of resources used (e.g., type of GPUs, internal cluster, or cloud provider)?
    \answerYes{The type of GPU is specified in \Cref{sec:experiments}.}
\end{enumerate}

\item If you are using existing assets (e.g., code, data, models) or curating/releasing new assets...
\begin{enumerate}
  \item If your work uses existing assets, did you cite the creators?
    \answerYes{Citations can be found in the descriptions of the experiments in  \Cref{sec:experiments}.}
  \item Did you mention the license of the assets?
    \answerYes{See the code provided as supplementary material.}
  \item Did you include any new assets either in the supplemental material or as a URL?
    \answerYes{The code, data and models are included in the supplemental material to reproduce the experiments.}
  \item Did you discuss whether and how consent was obtained from people whose data you're using/curating?
    \answerNo{The assets are licensed under the MIT License.}
  \item Did you discuss whether the data you are using/curating contains personally identifiable information or offensive content?
     \answerNA{}
\end{enumerate}

\item If you used crowdsourcing or conducted research with human subjects...
\begin{enumerate}
  \item Did you include the full text of instructions given to participants and screenshots, if applicable?
    \answerNA{}
  \item Did you describe any potential participant risks, with links to Institutional Review Board (IRB) approvals, if applicable?
      \answerNA{}
  \item Did you include the estimated hourly wage paid to participants and the total amount spent on participant compensation?
     \answerNA{}
\end{enumerate}

\end{enumerate}
\end{checklist}

%%%%%%%%%%%%%%%%%%%%%%%%%%%%%%%%%%%%%%%%%%%%%%%%%%%%%%%%%%%%

\newpage
\appendix

\section*{Appendix}

This appendix is organized as follows. In \Cref{sec:comp_spdes} we provide a summary of the computational aspects of SPDEs used for data simulation and model definition, emphasizing the important role of the Fourier Transform (\ref{ssec:Fourier}) for simulating noise realizations of Wiener processes (\ref{ssec:wiener}) and building numerical solvers for SPDEs (\ref{ssec:num_solvers}). In \Cref{sec:experiments_appendix} we provide additional considerations about our Neural SPDE model and further experimental details (\ref{ssec:experimental_details}) and additional experiments on the stochastic Ginzburg-Landau (\ref{ssec:GLE}) and wave (\ref{ssec:Wave}) equations, and on the deterministic Navier-Stokes PDE (\ref{ssec:det_PDEs}). 

\section{Computational aspects of SPDEs}\label{sec:comp_spdes}

We start this section with the definition of the \emph{Fourier Transform} (FT). We then define the \emph{Discrete Fourier Transform} (DFT) as an approximation to the FT of a function observed at finitely many locations. Next, we discuss the role played by the FT to sample realizations of Wiener processes, necessary to build spectral solvers for SPDEs. The interested reader is referred to \citet{briggs1995dft} and \citet{lord2014introduction} for further details. 

\subsection{The Fourier Transform}\label{ssec:Fourier}

% \subsubsection{Fourier Transform}\label{ssec:FT} 

Let $V$ be a vector space over the complex numbers (e.g. $\mathbb{C}^{d_h}$ or $\mathbb{C}^{d_h\times d_h}$). Let $r \in \mathbb{N}$ and let $\mathcal{C} \subset \mathbb{R}^r$ be a compact subset of $\mathbb{R}^r$. In the paper we used either $r=d$ and $\mathcal{C}=\mathcal{D}$ or $r=d+1$ and 
$\mathcal{C}=[0,T]\times\mathcal{D}$. 

\begin{definition}[\textbf{$r$-dimensional Fourier Transform}]\label{def:FT}
The $r$-dimensional FT $\mathcal{F}_r: L^2(\mathbb{R}^r, V) \to L^2(\mathbb{R}^r, V)$ and its inverse $ \mathcal{F}^{-1}_r: L^2(\mathbb{R}^r, V) \to L^2(\mathbb{R}^r, V)$ are defined as follows
\begin{align*}
    \mathcal{F}_r(f)(y) = \int_{\mathbb{R}^r} e^{-2\pi i \langle x,y \rangle}f(x)dx, \quad \mathcal{F}^{-1}_r(g)(x) = \int_{\mathbb{R}^r}e^{2\pi i \langle x , y \rangle}g(y)dy
\end{align*}
for any $f,g\in L^2(\mathbb{R}^r, V)$, where $i=\sqrt{-1}$ is the imaginary unit and $\langle \cdot, \cdot \rangle$ denotes the Euclidean inner product on $\mathbb{R}^r$. 
\end{definition}
% The Fourier transform enables to solve explicitly a large class of differential equations. FT, solve, inverse FT
% The (partial sum of the) Fourier series of a piecewise smooth periodic function $f$ converges pointwise for every $x$ to $(f(x^+)+f(x^-))/2$. At any point of continuity, the Fourier series converges to $f(x)$. Note that we can still define the Fourier coefficients for piecewise smooth functions compactly supported on $\mathcal{R}(L)$. If the Fourier series of $f$ converges on $\mathcal{R}(L)$, then it converges to the periodic extension of $f$ for all $x$, provided we use average values at points of
% discontinuity. 

In practice, we do not observe a function on $\mathbb{R}^r$ but on a subset $\mathcal{C}\subset\mathbb{R}^r$. Furthermore, functions are observed at finitely many locations in $\mathcal{C}$, and another transform---the discrete Fourier transform (DFT)---is used for numerical computations.

In the sequel we denote by $\Pi_N$ the set of periodic sequences indexed on $\mathbb{Z}_r$ with period vector $(N_1, \ldots, N_r)$.
\begin{definition}[\textbf{$r$-dimensional Discrete Fourier Transform}] The $r$-dimensional DFT $\mathcal{D}_r:\Pi_N \to \Pi_N$ and its inverse $\mathcal{D}_r^{-1}:\Pi_N \to \Pi_N$ are defined as follows,
\begin{align*}
    \mathcal{D}_r(u)_n= \sum_{k\in\mathbb{Z}^r\cap\mathcal{R}_N}u_ke^{-2\pi i \langle n,N^{-1}k\rangle},\quad
     \mathcal{D}^{-1}_r(v)_k= \frac{1}{|\mathrm{det} N|}\sum_{n\in\mathbb{Z}^r\cap\mathcal{R}_N}v_ne^{2\pi i \langle n,N^{-1}k\rangle}
\end{align*}
with $N=\mathrm{diag}(N_1, \ldots, N_r)\in\mathbb{N}^{r\times r}$, and $\mathcal{R}_N$ the rectangular domain $\mathcal{R}_N=\{x\in\mathbb{R}^r~|~0\leq x_i<N_i,~ i=1,\ldots,r\}$.
\end{definition} 
The DFT of a sequence can be computed exactly and efficiently using the \textit{fast Fourier transform} (FFT) algorithm \citep{cooley1965algorithm} which reduces the complexity from $\mathcal{O}(M^2)$ to $\mathcal{O}(M\log M)$ where $M=N_1N_2\ldots N_r$. Most importantly, the FFT algorithm is implemented in machine learning libraries such as PyTorch, which provide support for GPU acceleration and automatic differentiation capabilities.

Note that if we have a finite sequence, we may still define its DFT by implicitly extending the sequence periodically. In particular, when a compactly supported function is sampled on its interval of support, and the samples are used as input for a DFT, it is as if the periodic extension of  the function had been sampled. More precisely, consider an input sequence which corresponds to the evaluation of a function $f$ on a regular grid of $\mathcal{C}=\mathcal{R}_N$. For simplicity, suppose that $N_i=N_1$ for all $i=1,\ldots,r$ and consider the grid points $x_n=nL/N_1$ for $n\in\mathbb{Z}^r\cap\mathcal{R}_N$. Taking the DFT of the sequence of general term $u_n=f(x_n)$ we obtain for all $n\in\mathbb{Z}^r$, 
\begin{align*}
    \mathcal{D}_r(u)_n = \sum_{k\in\mathbb{Z}^r\cap\mathcal{R}_N}u_ke^{-2\pi i \langle n,k/N_1\rangle}
    = \sum_{k\in\mathbb{Z}^r\cap\mathcal{R}_N}f(x_k)e^{-2\pi i \langle y_n, x_k\rangle},
\end{align*}
where $y_{n}$ are the reciprocal frequency points given by $y_n=n/L$ for $n\in\mathbb{Z}^r\cap\mathcal{R}_N$.  The DFT of a compactly supported (or approximately compactly supported) function $f$ sampled on the regular grid of points $x_k$ approximates the FT of $f$ at the frequency points $y_n$ (up to a constant multiplicative factor).  

The FT is closely related to the notions of \textit{Fourier coefficients} and \textit{Fourier Series} defined hereafter.
\begin{definition}[\textbf{$r$-dimensional Fourier series}]
Let $f$ be a piecewise smooth function $f:\mathbb{R}^r\to V$ which is periodic in $x_i$ with period $L_i\in\mathbb{R}_+$ for all $i=1,\ldots,r$. The $r$-dimensional Fourier series of $f$ is a representation of the form,
\begin{align*}
    f(x) \sim \sum_{n\in\mathbb{Z}^r}c_n(f)e^{2\pi i\langle L^{-1}n, x\rangle},
\end{align*}
where $L=\mathrm{diag}(L_1, \ldots, L_r)\in\mathbb{R}^{r\times r}$ and $c_n(f)$ are complex coefficients, called \textit{Fourier coefficients}, given by 
\begin{align*}
    c_n(f) = \frac{1}{|\mathrm{det}L|}\int_{\mathcal{R}_L}e^{-2\pi i\langle L^{-1}n, x\rangle}f(x)dx, \ \ n\in\mathbb{Z}^r
\end{align*}
where $\mathcal{R}_L\subset \mathbb{R}^r$ denotes the rectangular domain of sides $L_1, \ldots, L_r$.
\end{definition}
We note that in the definition above, the sign $\sim$ means that the series is a formal series and no statement is made about the convergence of the series (the forms of convergence are studied in \citet{alimov1992multiple}). If $f$ is compactly supported on $\mathcal{R}_L$, we may still define its Fourier coefficients, and in this case $\mathcal{F}_r(f)(y_n)=|\mathrm{det} L|c_n(f)$ at the frequency points $y_n=L^{-1}n$.

\subsection*{Numerical consideration}
Consider a function $f$ which has compact support (or is periodic) which is observed at $M$ locations in its support (or its unitary cell $\mathcal{R}_L$). When using the DFT to approximate $M$ points of the spectrum $\mathcal{F}_r(f)(y_k)$ (or $M$ coefficients $c_k(f)$), a so-called \textit{aliasing} error usually occurs: due to the periodicity of the DFT, the $k^\text{th}$ coefficient of the DFT includes the contributions not only of the $k^\text{th}$ frequency mode, but also from higher modes of the underlying function $f$. In general the accuracy of the highest frequency modes is more impacted by this error, and aliasing occurs specifically when we compute nonlinear terms in the physical space. For example, in the main paper we approximate the evaluation on a discretization spatiotemporal grid $D\times\mathcal{T}$ of $\mathcal{F}_{d+1}^{-1}(\mathcal{F}_{d+1}(\mathcal{K})\mathcal{F}_{d+1}(f))$ by  $\mathcal{D}_{d+1}^{-1}(B\mathcal{D}_{d+1}(f|_{D\times\mathcal{T}}))$ where $f=\mathbbm{1}_{\geq 0}H_{\theta,\xi}(z)$ and $H_{\theta,\xi}$ is nonlinear. One possibility to mitigate aliasing is to set to zero the DFT terms (arising in nonlinearities) corresponding to the highest frequency modes before we apply the inverse DFT to go back to the physical space. This is precisely what we do when we parametrize only $k^1_{\text{max}}\times\ldots\times k^{d+1}_{\text{max}}\times d_h\times d_h$ entries of the complex tensor $B$, and set the others to zero, hence resolving potential aliasing errors. We note that specific rules have been proposed (notably in the literature on pseudo-spectral solvers) to deal with specific nonlinearities. However, in the context of Neural SPDE we learn the nonlinearities, hence the number of frequency modes that we retain is treated as an hyperparameter.

\subsection{Stochastic simulation of Wiener processes}\label{ssec:wiener}

After defining Wiener processes we outline the sampling procedure that we used to simulate the datasets in the main paper. For more details on computational aspects of SPDEs the reader is referred to \citet{lord2014introduction}. 

Throughout this section, $H$ will denote a separable Hilbert space (e.g. $H=L^2(\mathcal{D})$) with a complete orthonormal basis $\{\phi_k\}_{k \in \mathbb{N}}$. Let $(\Omega, \mathcal{F}, \mathcal{F}_t, \mathbb{P})$ be a filtered probability space.

\subsubsection{Q-Wiener process}

Consider an operator $\mathcal{Q} : H \to H$ such that there exists a bounded sequence of nonnegative real numbers $\{\lambda_k\}_{k \in \mathbb{N}}$ such that
$Q\phi_k = \lambda_k \phi_k$ for all $k\in\mathbb{N}$ (this is implied by $Q$ being a trace class, non-negative, symmetric operator, for example).

% \begin{definition}[\textbf{Trace class operator}] A non-negative definite operator $L\in\mathcal{L}(H)$ is of \emph{trace class} if~ $\mathrm{tr}(L)<\infty$ where the trace is defined by $\mathrm{tr}(L):=\sum_{j=1}^{\infty}\langle L\varphi_j,\varphi_j\rangle$, for an orthonormal basis $\{\varphi_j\}_{j\in\mathbb{N}}$ of $H$.
% \end{definition}

% \begin{definition}[\textbf{Covariance operator}] A linear
% operator $\mathcal{C}:H\to H$ is the covariance of the $H$-valued random variables $X$ and $Y$ if
% \begin{align*}
%     \langle\mathcal{C}\varphi, \psi\rangle = \mathrm{Cov}\left(\langle X,\varphi\rangle, \langle Y,\psi\rangle\right), \qquad \forall\varphi,\psi\in H
% \end{align*}
% \end{definition}
% If $\mathcal{C}$ is the zero operator, we say that $X$ and $Y$ are uncorrelated. 

% \begin{definition}[\textbf{$H$-valued Gaussian}] An $H$-valued random
% variable $X$ is Gaussian if $\langle X, \varphi\rangle$ is a real-valued Gaussian random variable for all $\varphi\in H$.
% \end{definition}

% \begin{proposition}
% Let $H$ be a separable Hilbert space and $X$ be an $H$-valued Gaussian with $\mu= \mathbb{E}[X]$. Then $X\in L^2(H)$ and the covariance operator 
% $\mathcal{C}$ of $X$ is a well-defined trace
% class operator. We write $X\sim\mathcal{N}(\mu,\mathcal{C})$.
% \end{proposition}

% \begin{proof}
% \citet[Corollary 4.41.]{lord2014introduction}.
% \end{proof}

\begin{definition}[\textbf{$Q$-Wiener process}] Let $Q$ be a trace class non negative, symmetric operator on $H$. A $H$-valued stochastic process $\{W(t):t\geq 0\}$ is called a $Q$-Wiener process if 
\begin{enumerate}
    \item $W(0)=0$ almost surely;
    \item $W(t;\omega)$ is a continuous sample trajectory $\mathbb{R}^+ \mapsto H$, for each $\omega \in \Omega$;
    \item $W(t)$ is $\mathcal{F}_t$-adapted and has independent increments $W(t)-W(s)$ for $s<t$;
    \item $W(t)-W(s)\sim\mathcal{N}(0,(t-s)Q)$ for all $0\leq s \leq t$.
\end{enumerate}
\end{definition}

In analogy to the Karhunen Lo\'eve expansion, it can be shown that $W(t)$ is a $Q$-Wiener process if and only if for all $t\geq 0$,
\begin{equation}\label{eqn:expansion_q_wiener}
    W(t)=\sum_{j=1}^{\infty}\sqrt{\lambda_j}\phi_j\beta_j(t)
\end{equation}
where $\beta_j(t)$ are i.i.d. Brownian motions, and the series converges in $L^2(\Omega,H)$. Moreover the series is $\mathbb{P}$-a.s. uniformly convergent on $[0, T ]$ for arbitrary $T > 0$. (i.e. converges in $L^2(\Omega, \mathcal{C}([0,T],H))$).

In the Navier-Stokes example, we drive the SPDE by samples $\xi$ from a $Q$-Wiener process in two dimensions. Here we follow \citet[Example 10.12]{lord2014introduction} and explain how the sampling procedure works in this case. Let $D=(0,L_1)\times(0,L_2)$ and consider an $L^2(D)$-valued Q-Wiener process $W(t)$. If the eigenfunctions of $Q$ are given by, $$\phi_k(x) = \frac{1}{\sqrt{L_1L_2}}e^{2i\pi(k_1x_1/L_1+ k_2x_2/L_2)}$$ numerical approximation of sample paths from $W(t)$ are easy to obtain through a DFT. Denote by $\lambda_k$ the eigenvalues of $Q$ (e.g. $\lambda_k = e^{-\alpha|k|^2}$ for some parameter $\alpha>0$) and let $\mathcal{J}$ be the index set defined by,  $$\mathcal{J}:=\{(j_1,j_2)\in\mathbb{Z}^2~:-J_1/2+1\leq j_1\leq J_1/2, ~-J_2/2+1\leq j_2\leq J_2/2\}$$
The goal is to sample from the truncated expansion of $W(t)$,
\begin{equation*}
    W^J(t) = \sum_{j\in\mathcal{J}}\sqrt{\lambda_{j}}\phi_{j}\beta_{j}(t),
\end{equation*} 
at the collection of sample points,
\begin{equation*}
     x_k=\left(L_1k_1/J_1, L_2k_2/J_2\right)^T, \qquad 0\leq k_1\leq J_1-1,~0\leq k_2\leq J_2-1.
\end{equation*}
Consider the random variable $Z(t_n,x)$ defined by, 
\begin{equation*}
    Z(t_n,x)= \sqrt{\Delta t} \sum_{j\in\mathcal{J}}\sqrt{\lambda_{j}}\phi_{j}(x)\xi^n_{j}, \qquad \xi^n_{j}\sim \mathbb{C}\mathcal{N}(0,2),
\end{equation*}
meaning that $\xi^n_j=a+ib$ with $a,b\overset{\text{i.i.d}}{\sim}\mathcal{N}(0,1)$ such that $Z(t_n,x_{k})$ is a complex random variable with independent real and imaginary part with the same distribution as two independent copies of the increment $W^J(t_n+\Delta t,x_{k})-W^J(t_n,x_{k})$. Furthermore, $Z(t_n,x_k)$ can be expressed in the form, 
\begin{align}\label{eq:increment2}
  Z(t_n,x_k) &= \frac{1}{J_1J_2} \sum_{j_1=-J_1/2+1}^{J_1/2}\sum_{j_2=-J_2/2+1}^{J_2/2}\widetilde{Z}_{j_1,j_2}e^{2i\pi\left(j_1\frac{k_1}{J_1}+j_2\frac{k_2}{J_2}\right)}
\end{align}
where $\widetilde{Z}_{j_1,j_2}=\sqrt{\Delta t \lambda_{j_1,j_2}}J_1J_2\xi^n_{j_1,j_2}$
We recognize that the matrix with entries given by \cref{eq:increment2} is the 2D inverse DFT of the $J_1\times J_2$ matrix with entries $\widetilde{Z}_{j_1,j_2}$. Therefore, we can sample two independent copies of 
\begin{align*}
    W^J(t_n+\Delta t,x_{k})-W^J(t_n,x_{k}), \ \ 0\leq k_1\leq J_1-1,~0\leq k_2\leq J_2-1
\end{align*}
by computing a single 2D inverse DFT.

\subsubsection{Cylindrical Wiener process}

If the operator $Q=I$ is the identity, then $Q$ is not of trace class on $H$ so that the series in \cref{eqn:expansion_q_wiener} does not converge in $L^2(\Omega,H)$. This motivates the definition of cylindrical Wiener processes.

\begin{definition}[\textbf{Cylindrical Wiener process}] Let $H$ be a separable Hilbert space. A cylindrical Wiener process (a.k.a space-time white noise) is a $H$-valued stochastic process $\{W(t):t\geq 0\}$ defined by  
\begin{align}\label{eqn:expansion_cyl_wiener}
    W(t) = \sum_{j=1}^{\infty}\phi_j\beta_j(t)
\end{align}
where $\{\phi_j\}$ is any orthonormal basis of $H$ and $\beta_j(t)$ are i.i.d. Brownian motions.
\end{definition}

In all examples except Navier-Stokes, we drive the SPDE by samples $\xi$ from a cylindrical Wiener process in one dimension. Let $D=(0,L)$ and consider an $L^2(D)$-valued cylindrical Wiener process $W(t)$. As explained in \citet[Example 10.31]{lord2014introduction}, if we take the basis $$\phi_k(x) = \sqrt{2/L}\sin{(k\pi x/L)}$$ numerical approximation of sample paths from $W(t)$ are easy to obtain. The goal is to sample from the truncated expansion,
\begin{align}\label{eq:projected_W}
    W^J(t) = \sum_{j=1}^{J}\phi_{j}\beta_{j}(t),
\end{align} at the collection of sample points $x_k=kL/J$ for $k=1,\ldots,J$. Observing that a trigonometric identity yields,
\begin{align*}
    \mathrm{Cov}\left(W^J(t,x_i),W^J(t,x_k)\right) = (tL/J)\delta_{ik}, \qquad i,k=1,\ldots,J
\end{align*}
the increments $W^J(t_n+\Delta t,x_k)-W^J(t_n,x_k)\sim\mathcal{N}(0,\Delta tL/J)$ for all $k=1,\ldots,k$.

\subsection{Numerical solvers}\label{ssec:num_solvers}

In this section we present an overview of the numerical solvers for SPDEs we used to generate the data for all the experiments. The stochastic Ginzburg-Landau (\Cref{ssec:GL,ssec:GLE}), stochastic wave (\Cref{ssec:Wave}) equations have been solved using the finite difference method, while the stochastic Korteweg–De Vries (\Cref{ssec:KdV}) and Navier Stokes (\Cref{ssec:SNS}) equations have been solved using the spectral Galerkin method. We use the same setup as in \Cref{sec:background}. In particular, we focus on stochastic semilinear evolution equations of the form
\begin{equation}\label{eqn:spde_appendix}
    du_t = \left(\mathcal{L}u_t + F(u_t)\right)dt + G(u_t)dW_t
\end{equation}
where $W_t$ is either a $Q$-Wiener process or a cylindrical Wiener process and $\mathcal{L}$ is a linear differential operator generating a semigroup $e^{t\mathcal{L}}$. We consider nonlinearities $F,G$ regular enough (see \citet[Assumption 10.23]{lord2014introduction}) to guarantee existence and uniqueness of mild solutions of \cref{eqn:spde_appendix} \citep[Thm. 10.26]{lord2014introduction}.

\subsubsection{Finite difference method}

We illustrate this numerical method for the reaction-diffusion equation
\begin{equation*}
    du_t = \left(\epsilon \partial^2_{xx} u + F(u_t)\right)dt + \sigma dW_t, \quad u(0,x) = u_0(x),
\end{equation*}
with homogeneous Dirichlet boundary conditions and where $\epsilon,\sigma>0$ are constants. We assume for simplicity that $u_0,u_t,W_t$ are real-valued and $\mathcal{D}=(0,a)$. The generalization to higher dimensions is straightforward.

Consider the grid points $x_j = jh$, where $h=\frac{a}{J}$ and $j=0,...,J$, for some spatial resolution $J \in \mathbb{N}$. Let $u_J(t)$ be the \emph{finite difference approximation} of $[u(t,x_1),...,u(t,x_{J-1})]$ (similarly for $W_J(t)$) resulting from the solution of the following SDE
\begin{equation*}
    du_J(t) = [-\epsilon M u_J(t) + \hat f(u_J(t))]dt + \sigma dW_J(t)
\end{equation*}
where $\hat f(u_J) = [f(u_1),...,f(u_{J-1})]^T$ and $M$ is the $(J-1) \times (J-1)$ matrix approximating Laplacian (with free boundary conditions) which is given by
\begin{equation*}
    M = \frac{1}{h^2}\begin{pmatrix}
                        2 & -1 &  &  &  &    \\
                       -1 & 2 & -1 &  &  &      \\
                         & -1 & 2 & -1 &  &    \\
                         & & \ddots & \ddots & \ddots &  \\ 
                         &  &  & -1 & 2 & -1  \\
                         &  & &  & -1 & 2
                      \end{pmatrix}
\end{equation*}
One could modify $M$ for specific boundary conditions. For instance in the case of periodic boundary one should modify $M_{1,J-1} = M_{J-1, 1} = -1$ (see \citet[Chapter 3.4]{lord2014introduction} for Dirichlet and Neuman boundary condition modifications of $M$).
To discretize in time, we may apply numerical methods for SDEs (see for example \citet[Chapter 8]{lord2014introduction}). Choosing the standard Euler-Marayama scheme with time step $\Delta t$ yields an approximation $u_{J,n}$ to $u_J(t_n)$ at $t_n = n \Delta t$ defined by
\begin{equation*}
    u_{J,n+1} = (I + \Delta t \epsilon M)^{-1}\left(u_{J,n} +  \hat f(u_{J,n})\Delta t + \sigma (W_J(t_{n+1}) - W_J(t_{n}))\right)
\end{equation*}
The increments $(W_J(t_{n+1}) - W_J(t_{n}))$ are generated using techniques discussed in \Cref{ssec:wiener}.

\subsubsection{Spectral Galerkin method}

Consider again a separable Hilbert space $H$. Assume that the differential operator $\mathcal{L}$ in \cref{eqn:spde_appendix} has a complete set of orthonormal eigenfunctions $\{\phi_j\}_{j \in \mathbb{N}}$ and eigenvalues $\lambda_j<0$, ordered so that $\lambda_{j+1}<\lambda_j$. Then, we can define the semigroup $e^{t\mathcal{L}}$ as follows
\begin{equation*}
    e^{t\mathcal{L}}h = \sum_{j=1}^\infty e^{\lambda_jt}\langle h, \phi_j \rangle \phi_j, \quad h \in H.
\end{equation*}
Define the \emph{Galerkin subspace} $V_J = \text{Span}\{\phi_1,...,\phi_J\}$ and the orthonormal projections $P_J:H\to V_J$ as follows
\begin{equation*}
    P_J h = \sum_{i=1}^J \langle u, \phi_j \rangle \phi_j, \quad h \in H.
\end{equation*}
Then, the following defines \emph{spectral Galerkin approximation} of \cref{eqn:spde_appendix}
\begin{equation*}
    d u_J(t) = (\mathcal{L}_J u_J(t) + P_J F(u_J(t)))dt + P_JG(u_J(t))dW_J(t), \quad u_J(0) = P_Ju_0
\end{equation*}
where $u_J := P_Ju$ and $\mathcal{L}_J:=P_J\mathcal{L}$ and $W_J = P_J W$ is as in~\eqref{eq:projected_W}. Using a Euluer-Marayama discretization as above, we obtain the following discretization
\begin{equation*}
    u_{J,n+1} = (I + \Delta t \mathcal{L}_J)^{-1}(u_{J,n} + \Delta t P_J F(u_{J,n}) + P_J G(u_{J,n})\Delta W_{J,n}).
\end{equation*}
This approach is particularly convenient for problems with additive noise where the eigenfunctions of $\mathcal{L}$ and $Q$ (the covariance of the $Q$-Wiener process $W$) are equal, which is the case for all the experiments in this paper generated with this method. The eigenfunctions of the Laplacian with periodic boundary conditions correspond to the Fourier basis exponentials; therefore, one can define the projection $P_J$ in terms of the DFT.

\section{Further experiments}\label{sec:experiments_appendix}
In this section with discuss additional details about the NSPDE model, its training procedure an of the baseline models, including how the relevant hyperparameters have been selected for each model.

\subsection{Derivation of the ODE parameterisation}\label{sec:appendix_ODE_derivation}

If one assumes that $\mathcal{L}$ is a polynomial differential operator of degree $N$ of the form
\begin{align*}
    \mathcal{L} = \sum_{n=0}^N \sum_{\substack{n_1,...,n_d \\ n_1+...+n_d=n}}C_{n_1,...,n_d}\frac{\partial^n}{\partial x_1^{n_1}...\partial x_d^{n_d}},
\end{align*}
where $C_{n_1,...,n_d} \in \mathbb{C}^{d_h \times d_h}$ are complex matrices, then the FT of the kernel associated to $\mathcal{L}$ satisfies
\begin{equation*}
    \mathcal{F}(\mathcal{K}_t)(y) = e^{tP(iy)} \in \mathbb{C}^{d_h \times d_h},
\end{equation*}
for any $y \in \mathbb{C}^d$, where $e$ is the matrix exponential and $P$ is the following matrix-valued polynomial
\begin{equation*}
    P(y) = \sum_{n=0}^N \sum_{\substack{n_1,...,n_d \\ n_1+...+n_d=n}} (2\pi)^n  y_1^{k_1}...y_d^{k_d}C_{n_1,...,n_d}.
\end{equation*}
Therefore, there exists a map $A :  \mathbb{C}^d \to \mathbb{C}^{d_h \times d_h}$ such that $\mathcal{F}(\mathcal{K}_{t})(y) = e^{tA(y)}$. It follows that 
\begin{equation*}
    z_t = \mathcal{F}^{-1}\Big(e^{tA} \mathcal{F}(z_0) + \int_0^t e^{(t-s)A}\mathcal{F}(H_{\theta,\xi}(z_s))ds\Big) = \mathcal{F}^{-1}(v_t),
\end{equation*}
where $v_t : \mathbb{C}^d \to \mathbb{C}^{d_h}$ is the solution of the following ODE
\begin{align*}
    v_t = v_0 + \int_0^t Av_s + \mathcal{F}(H_{\theta,\xi}(\mathcal{F}^{-1}(v_s))).
\end{align*}
as shown in section 3.2.

\subsection{Additional experimental details}\label{ssec:experimental_details}

For all experiments the dataset is split into a training, validation and test sets with relative sizes $70\%/15\%/15\%$. For all models, a grid search on the hyperparameters is performed using the training and validation sets. We use the Adam optimizer and a scheduler which reads the validation loss and reduces the learning rate if no improvement is seen for a \textit{patience} number of epochs. Additionally, an early stopping method is used to halt the training of the model if no improvement is seen after a \textit{patience} number of epochs. The hyperparameters included in the grid search are stated below and examples of hyperparameter selection results are provided in \crefrange{table:ncde_fno_hyper}{table:nspde_hyper}.

\paragraph{NSPDE} The hyperparameters included in the grid search are the number of frequency modes used to parametrize the kernel in Fourier space $B=\mathcal{F}_{d+1}(\mathcal{K})$ and the number of forward iterations used to solve the fixed point problem. 

\paragraph{FNO} The hyperparameters included in the grid search are the number of frequency modes used to parametrize the kernel and the number of layers $M$. Note that the numbers of frequency modes in the grid search differ from the ones used for the NSPDE model by a factor $2$ to ensure that the effective number of retained modes is the same. For both the NSPDE model and FNO, we kept the number of hidden channels fixed to $d_h=32$ as this systematically yielded better performances than previously included values and enabled to perform the grid search in a reasonable time. 

\paragraph{DeepONet}
The \emph{Deep Operator Network} (DeepONet) \citep{lu2021learning} is another popular class of neural network models for learning operators on function spaces. The DeepONet architecture is based on the universal approximation theorem of \citet{chen1995universal}. It consists of two sub-networks referred to as the \emph{branch} and the \emph{trunk} networks. The trunk acts on the coordinates $(t,x)\in[0,T]\times\mathcal{D}$, while the branch acts on the evaluation of the initial condition $u_0$ on a discretized grid $D$. Therefore, the DeepONet is not a space resolution-invariant architecture. The output of the network is expressed as $$\text{DeepONet}(u_0)(t,x)=\sum_{k=1}^{p}b_k(u_0)\tau_k(t,x)+b_0,$$
where the $b_k$ and the $\tau_k$ are the outputs of the branch and trunk network respectively. The trunk network is usually a feedforward neural network, and one can chose the architecture of the branch network depending on the structure of the input domain. We follow \citet{lu2021learning} and use feedforward neural networks for both the trunk and the branch networks. We perform a grid search on the depth and width of the trunk and branch feedforward neural networks. 

\paragraph{NRDE/NCDE/NCDE-FNO} The hyperparameters included in the grid search are the number of hidden channels and the type  of solver as implemented by $\text{torchdiffeq}$ \citep{chen2018neural}. We note that we used a depth-2 NRDE model (depth-2 already results in $d_\xi=8\,385$ for forcings observed at $128$ spatial points  and higher depths models could not fit in memory) and recall that NCDE is a depth-1 NRDE.

\begin{table}[H]

\sisetup{round-mode=places, round-precision=3}
 \begin{minipage}{.50\linewidth}
    \caption{Grid search NCDE (KdV)}
     \centering
     \vspace{5pt}
    \begin{tabular}{cccc}%
    \toprule
    \bfseries $d_h$ & \# parameters & solver & validation loss % specify table head
    \begin{filecontents*}{fig/log_ncde_kdv_xi_05_1.csv}
dh,nbparams,losstrain,lossval,losstest,method
8,136464,0.481814626,0.510894516,0.467468876,rk4
16,272672,0.479413868,0.509518019,0.466594505,rk4
32,545088,0.47678406,0.505248965,0.463742028,rk4
8,136464,0.521419986,0.559837013,0.509130245,euler 
16,272672,0.513829332,0.56092666,0.511447265,euler 
32,545088,0.511523652,0.55550838,0.505043851,euler 
\end{filecontents*}
\csvreader[head to column names]{fig/log_ncde_kdv_xi_05_1.csv}{}% use head of csv as column names
    {\\\hline\ \dh & \nbparams & \method & \num{\lossval}}\\% specify your coloumns here
    \bottomrule
    \end{tabular}%    
    \label{table:ncde_fno_hyper}
\end{minipage}
 \begin{minipage}{.50\linewidth}
    \caption{Grid search NCDE-FNO (KdV)}
     \centering
      \vspace{5pt}
    \begin{tabular}{cccc}%
    \toprule
    \bfseries $d_h$ & \# parameters & solver & validation loss % specify table head
    \begin{filecontents*}{fig/log_ncdefno_kdv_xi_05_1.csv}
dh,nbparams,losstrain,lossval,losstest,method
8,5761,0.129483202,0.139682349,0.125816479,rk4
16,15617,0.130981224,0.141644616,0.127252935,rk4
32,48769,0.134059668,0.144638685,0.130473651,rk4
8,5761,0.290530209,0.310232194,0.2827158,euler
16,15617,0.293410838,0.314498353,0.283325545,euler
32,48769,0.298999404,0.320776007,0.289936108,euler
\end{filecontents*}
\csvreader[head to column names]{fig/log_ncdefno_kdv_xi_05_1.csv}{}% use head of csv as column names
    {\\\hline\ \dh & \nbparams & \method & \num{\lossval}}\\% specify your coloumns here
    \bottomrule
    \end{tabular}%
\end{minipage}
\end{table}
% \begin{table}[H]
%     \caption{Grid search NRDE (KdV)}
%      \centering
%      \vspace{5pt}
%     \begin{tabular}{cccc}%
%     \toprule
%     \bfseries $d_h$ & \# parameters & solver & validation loss % specify table head
%     \csvreader[head to column names]{fig/log_ncde_kdv_xi_05_1.csv}{}% use head of csv as column names
%     {\\\hline\ \dh & \nbparams & \method & \num{\lossval}}\\% specify your coloumns here
%     \bottomrule
%     \end{tabular}%    
%     \label{table:ncde_fno_hyper}
% \end{table}

\begin{table}[H]

\sisetup{round-mode=places, round-precision=3}
    \caption{Grid search DeepONet (KdV)}
     \centering
     \vspace{5pt}
    \begin{tabular}{ccccc}%
    \toprule
    Branch \& trunk width & Branch depth & Trunk depth  & \# parameters & validation loss % specify table head
    \begin{filecontents*}{fig/log_deeponet_kdv_xi_05_1.csv}
width,bd,td,nbparams,losstrain,lossval,losstest
256,3,4,1935616,0.221399336,0.258433917,0.235080322
128,3,4,885888,0.229563555,0.269294418,0.246182195
128,3,3,869376,0.246909998,0.278619817,0.256484305
128,3,2,852864,0.240199801,0.284010718,0.261887156
512,3,4,4526592,0.259697148,0.293649146,0.266763306
512,2,2,3738624,0.207179598,0.294863807,0.274834765
256,4,4,2001408,0.239479885,0.29493196,0.270941454
512,4,4,4789248,0.24883233,0.30195588,0.272800165
256,2,2,1738240,0.212313177,0.30448823,0.284467795
256,3,3,1869824,0.280873673,0.306676544,0.283472305
128,4,3,885888,0.249249223,0.310608045,0.283307785
128,4,2,869376,0.268621668,0.310997648,0.28643318
256,3,2,1804032,0.287349082,0.312803613,0.289982846
128,2,2,836352,0.2164334,0.31450728,0.286477428
256,4,2,1869824,0.273010915,0.315795361,0.293194064
128,4,4,902400,0.248809069,0.318009085,0.297533499
256,4,3,1935616,0.269465939,0.318738336,0.289773276
512,4,3,4526592,0.274002869,0.319533025,0.298067464
512,4,2,4263936,0.300889079,0.333814054,0.310087956
512,3,2,4001280,0.264544799,0.349760381,0.31907501
128,2,4,869376,0.187322027,0.349763536,0.329715824
512,3,3,4263936,0.324112919,0.359194387,0.323586655
512,2,3,4001280,0.271546238,0.36570065,0.34149327
256,2,3,1804032,0.270658441,0.370141753,0.350101855
128,2,3,852864,0.255744425,0.381786492,0.365174029
256,2,4,1869824,0.239978407,0.391814762,0.363569699
512,2,4,4263936,0.278807229,0.395355135,0.376916546
\end{filecontents*}
\csvreader[head to column names]{fig/log_deeponet_kdv_xi_05_1.csv}{}% use head of csv as column names
    {\\\hline\ \width & \bd & \td & \nbparams & \num{\lossval}}\\% specify your coloumns here
    \bottomrule
    \end{tabular}
    \label{table:deeponet_hyper}
\end{table}

\begin{table}[H]

\sisetup{round-mode=places, round-precision=3}
    \caption{Grid search FNO (KdV)}
     \centering
     \vspace{5pt}
    \begin{tabular}{cccccc}%
    \toprule
    \bfseries $d_h$ & depth & modes 1 & modes 2 & \# parameters & validation loss % specify table head
    \begin{filecontents*}{fig/log_fno_kdv_xi_1_1.csv}
dh,L,modesx,modesy,nbparams,losstrain,lossval,losstest
32,1,16,16,532993,0.156995809,0.163146141,0.152758058
32,1,16,50,1647105,0.103278269,0.117178023,0.111896586
32,1,32,16,1057281,0.157165097,0.162998329,0.152589499
32,1,32,50,3285505,0.104487291,0.117385987,0.112118402
32,2,16,16,1058337,0.156777787,0.162687706,0.152253383
32,2,16,50,3286561,0.105204234,0.119699164,0.114808177
32,2,32,16,2106913,0.156736134,0.162669309,0.152263639
32,2,32,50,6563361,0.098062306,0.118064997,0.113626681
32,3,16,16,1583681,0.156873809,0.162573538,0.152147375
32,3,16,50,4926017,0.108174823,0.120414368,0.115606752
32,3,32,16,3156545,0.156884619,0.162586942,0.152146921
32,3,32,50,9841217,0.096742274,0.120086302,0.11499597
32,4,16,16,2109025,0.156802443,0.162558194,0.152151839
32,4,16,50,6565473,0.107600884,0.121534047,0.116632679
32,4,32,16,4206177,0.156854136,0.162577159,0.152141147
32,4,32,50,13119073,0.097466433,0.120293992,0.115513122
\end{filecontents*}
\csvreader[head to column names]{fig/log_fno_kdv_xi_1_1.csv}{}% use head of csv as column names
    {\\\hline\ \dh & \L & \modesx & \modesy & \nbparams & \num{\lossval}}\\% specify your coloumns here
    \bottomrule
    \end{tabular}
    \label{table:fno_hyper}
\end{table}

\begin{table}[H]

\sisetup{round-mode=places, round-precision=3}
    \caption{Grid search NSPDE (KdV)}
     \centering
     \vspace{5pt}
    \begin{tabular}{cccccc}%
    \toprule
    \bfseries $d_h$ & Picard's iterations & modes 1 & modes 2 & \# parameters & validation loss % specify table head
    \begin{filecontents*}{fig/log_nspde_kdv_xi_1_1.csv}
dh,iter,modesx,modesy,nbparams,losstrain,lossval,losstest
32,1,32,32,1055233,0.019290659,0.022924113,0.021813703
32,1,32,100,3283457,0.00978041,0.011412465,0.011213234
32,1,64,32,2103809,0.02498228,0.029829686,0.027923083
32,1,64,100,6560257,0.008375205,0.009496049,0.009342044
32,2,32,32,1055233,0.013751543,0.017592222,0.016827066
32,2,32,100,3283457,0.008716699,0.011650713,0.011194938
32,2,64,32,2103809,0.012566753,0.015297543,0.014715799
32,2,64,100,6560257,0.008259323,0.01000622,0.009838308
32,3,32,32,1055233,0.012086184,0.016452566,0.015959209
32,3,32,100,3283457,0.010705066,0.013158875,0.013256771
32,3,64,32,2103809,0.01760905,0.021904041,0.02011299
32,3,64,100,6560257,0.009017411,0.011370487,0.011274823
32,4,32,32,1055233,0.013977978,0.018861536,0.01854095
32,4,32,100,3283457,0.0091733,0.011867433,0.01145267
32,4,64,32,2103809,0.016786256,0.020876131,0.020390217
32,4,64,100,6560257,0.013415584,0.016275374,0.016218475
\end{filecontents*}
\csvreader[head to column names]{fig/log_nspde_kdv_xi_1_1.csv}{}% use head of csv as column names
    {\\\hline\ \dh & \iter & \modesx & \modesy & \nbparams & \num{\lossval}}\\% specify your coloumns here
    \bottomrule
    \end{tabular}
    \label{table:nspde_hyper}
\end{table}

\subsection{Stochastic Ginzburg-Landau equation}\label{ssec:GLE}

Recall that the stochastic Ginzburg-Landau equations are of the form,
\begin{align}
    \partial_t u - \Delta u &= 3u -u^3 + G(u)\xi,  \\
    u(0,x) &= u_0(x), \quad (t,x)\in [0,T]\times[0,1] \nonumber
\end{align}

subject to either Periodic or Dirichlet boundary conditions. Periodic boundary conditions are given by $u(t,0) = u(t,1)$ for all $t \geq 0$ and Dirichlet boundary conditions are given by $u(t,0) = u(t,1) = 0$ for all $t \geq 0$. Initial condition we take as in \Cref{ssec:GL} $u_0(x) = x(1-x) + \kappa \eta(x)$ with $\kappa = 0$ or $\kappa = 0.1$ depending on a task. In both Periodic and Dirichlet case we can take $\eta(x)$ as in \Cref{ssec:GL} though in Dirichlet case one must take $a_0 = 0$ to ensure $u_0$ being zero at the boundary.

We first reproduce an experiment from \Cref{ssec:GL} on the additive stochastic Ginzburg-Landau equation but with Dirichlet boundary conditions instead of the periodic. We compare it to the benchmark of FNO model which was the most successful among all the benchmarks of \Cref{sec:experiments}. From \Cref{table:phi41d} we see that even though Neural SPDE model depends on the spectral methods the errors did not increase compared to the periodic equation in \Cref{ssec:GL} (see \Cref{table:phi41}). Our algorithm still outperforms FNO whose relative $L2$ error increased slightly. The fact that Neural SPDE can be applied to non-periodic equations could be perhaps explained by interpolation ($L_\theta$) and projection ($\Pi_\theta$) neural networks that could correct for non-periodicity of the data.

\begin{table}[H]
\caption{\textbf{Additive stochastic Ginzburg-Landau equation with homogeneous Dirichlet boundary conditions}. The experimental setup is the same as in the main paper. We report the relative L2 error on the test set. The symbol x indicates that the model is not applicable. $N$ is fixed to $1\,000$.}
\begin{center}
\begin{tabular}{lccc}
\toprule
  Model & $u_0\mapsto u$ & $\xi\mapsto u$ & $(u_0,\xi)\mapsto u$ \\
\midrule 
FNO & 0.132 & 0.023 & x \\
\midrule
NSPDE (Ours)  & 0.135 & 0.008 & 0.010 \\
\bottomrule
\end{tabular}
\label{table:phi41d}
\end{center}
\end{table}

We now take a look at the specific hyperparameter: number of forward iterations in the fixed point solver. We also call this a number of Picard iterations $P$. Theoretically as $P$ increases Fixed Point Solver should converge to the true solution (see \citep{hairer2009introduction}). This suggests that higher $P$ should improve the performance of the Neural SPDE algorithms. In practise we observed in both additive Ginsburg Landau equation from \Cref{ssec:GL} and in KdV equation from \Cref{ssec:KdV} that $P = 1$ could already be enough. This could be explained either by dominance of the linear part of the equation or by overfitting in these cases. Thus we present an experiment on the multiplicative stochastic Ginzbug-Landau equation over a longer (compared to \Cref{ssec:GL}) time interval. In the \Cref{table:phi41m} we compare NSPDE with $P \in \{1,2,3,4\}$ and again include FNO benchmark (which performed best in the previous experiments). We see that NSPDE with even $P = 1$ outperforms FNO. Relative $L2$ error for $T = 0.05$ increases for both NSPDE and FNO due to more complicated multiplicative noise. In \Cref{table:phi41m} we present for each $P$ the best result over other hyperparameters obtained by cross validation. One could clearly see an improvement in error as we increase the number of Picard iterations $P$ (with an exception of the case $T = 0.05$ where $P = 3$ outperformed $P = 4$). This improvement becomes more apparent as the time frame $T$ increases. Heuristically (and qualitatively) this is due to the fact that for the short times solution of the SPDE is relatively close to its linearised version and that nonlinearity of the equation starts to play a bigger role for larger $T$.

\begin{table}[H]
\caption{\textbf{Multiplicative stochastic Ginzburg-Landau equation}. We report the relative L2 error on the test for FNO and NSPDE (Ours) for different number of Picard iterations on the task $\xi\to u$.}
\begin{center}
\begin{tabular}{l|ccccc}
\toprule
 Time horizon & FNO & NSPDE ($P=1$) & NSPDE ($P=2$) & NSPDE ($P=3$) & NSPDE ($P=4$) \\
 \midrule
$T=0.05$ & 0.040 & 0.023 & 0.018 & \textbf{0.016} & 0.017 \\
$T=0.10$ & 0.068 & 0.042 & 0.041 & \textbf{0.040} & \textbf{0.040} \\
$T=0.25$ & 0.105 & 0.079 & 0.077 & 0.073 & \textbf{0.072} \\
\bottomrule
\end{tabular}
\label{table:phi41m}
\end{center}
\end{table}

\subsection{The stochastic wave equation}\label{ssec:Wave}
In this section we consider the following nonlinear wave equation with multiplicative stochastic forcing,
\begin{align}\label{eqn:wave}
    \partial_t^2 u - \Delta u &= \cos(\pi u) +u^2 + u\xi,\\
    u(t,0) &= u(t,1), \nonumber \\
    % \partial_t u(0,x) &= x(1-x), \nonumber\\
    u(0,x) &= u_0(x), \nonumber \\
    \partial_t u(0,x) &= v_0(x), \quad (t,x)\in[0,T]\times[0,1].\nonumber
\end{align}
The nonlinear stochastic wave equation arises in relativistic quantum mechanics and is also used in simulations of nonlinear waves that are subject to either noisy observations or random forcing. We refer a reader to~\citet{temam2012infinite} for an overview on the nonlinear wave equation. The above equation can put in a form of \cref{eqn:SPDE} by rewriting it as a system for $(u,v) = (u, \partial_t u)$. To generate training datasets, we solve the SPDE using a finite difference method with $128$ evenly distanced points in space and a time step size $\Delta t=10^{-3}$. As in \citet[eq. (3.5)]{chevyrev2021feature}, we solve the SPDE until $T=0.5$. We then downsample the temporal resolution by a factor $5$, resulting in $100$ time points. Here, the initial condition is given by $u_0(x)=\sin(2\pi x)+\kappa\eta(x)$, where $\eta$ is defined in \Cref{ssec:GL} and for simplicity initial velocity $v_0$ is taken deterministic $v_0(x) = x(1-x)$. Similarly to \Cref{ssec:GL} we either take $\kappa=0$ or $\kappa=1$ to generate datasets where the initial condition is either fixed or varies across samples. Each dataset consists of $N=1\,000$ training observations.

\begin{table}[H]
\caption{\textbf{Stochastic Wave equation}. We report the relative L2 error on the test set. The symbol x indicates that the model is not applicable. $N$ is fixed to $1\,000$.}
\begin{center}
\begin{tabular}{lccc}
\toprule
  Model & $u_0\mapsto u$ & $\xi\mapsto u$ & $(u_0,\xi)\mapsto u$ \\
\midrule 
NCDE & x & 0.142  & 0.432 \\
NRDE & x & 0.146 & 0.445 \\
NCDE-FNO & x & 0.029 & 0.037  \\
DeepONet & 0.190 & 0.143 & x \\
FNO & 0.151 & 0.026 & x \\
\midrule
NSPDE (Ours)  & 0.150 & \textbf{0.023} & \textbf{0.026} \\
\bottomrule
\end{tabular}
\label{table:wave}
\end{center}
\end{table}

\subsection{Deterministic Navier-Stokes PDE}\label{ssec:det_PDEs}

\begin{figure*}[h]
    \centering
    \includegraphics[scale=0.25]{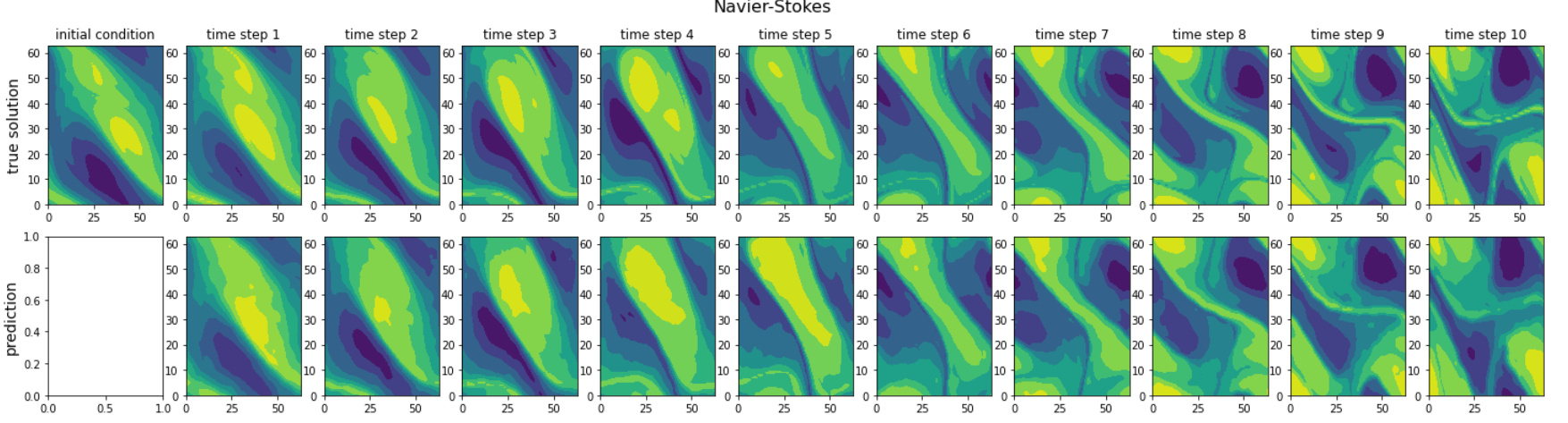}
    \caption{\textbf{Top panel:} Initial vorticity and ground truth vorticity at later time steps on a $64\times64$ mesh. \textbf{Bottom panel:} Predictions of the Neural SPDE model.}
    \label{fig:my_label}
\end{figure*}

In this final experiment, we demonstrate that our Neural SPDE model can also be used in the setting of PDEs without any stochastic term. We do so by studying the example from \cite{li2020fourier} on deterministic Navier-Stokes. More precisely, we consider the 2D Navier-Stokes equation  for a viscous, incompressible fluid in vorticity form:
\begin{align}
    \partial_t w(t,x) - \nu \Delta w(t,x) &= f(x) - u(t,x) \cdot \nabla w(t,x), && t \in [0,T], x \in [0,1]^2 \\
    \nabla \cdot u(t,x) &= 0, && t \in [0,T], x \in [0,1]^2 \\
    w(x,0) &= w_0(x), && x \in [0,1]^2
\end{align}

where $u : [0,T] \times [0,1]^2 \to \mathbb{R}^2$ is the velocity field, $w = \nabla \times u$ is the vorticity with $w_0 : [0,1]^2 \to \mathbb{R}$ being the initial vorticity. Here $f$ is a deterministic forcing term which we take as in \cite{li2020fourier}.  We follow the experimental setup from \cite{li2020fourier} and use the dataset (available under an MIT license) where $\nu=10^{-5}$, $N=1000$ and $T=20$. We achieve similar performances as FNO with a L2 error of $0.17$. A comparison between a true and predicted trajectory is depicted in \Cref{fig:my_label}.

\end{document}